\documentclass[journal]{IEEEtran}

\usepackage[utf8]{inputenc}
\usepackage[T1]{fontenc}

\usepackage{bm}
\usepackage[cmex10]{amsmath}
\usepackage{amsfonts}
\usepackage{amsthm}

\usepackage{graphicx}
\usepackage{subcaption}
\usepackage{fixltx2e}

\usepackage{cite}

\usepackage{url}

\usepackage{booktabs}
\usepackage{xcolor}
\usepackage{multirow}

\usepackage{color}

\usepackage{listings}
\usepackage{algorithm}
\usepackage{algorithmic}

\definecolor{deepblue}{rgb}{0,0,0.5}
\definecolor{deepred}{rgb}{0.6,0,0}
\definecolor{deepgreen}{rgb}{0,0.5,0}
\makeatletter
\def\lst@makecaption{%
  \def\@captype{table}%
  \@makecaption
}
\makeatother

\newcommand{\mR}{\mathbb{R}}
\newcommand{\mnorm}[1]{\left\lVert#1\right\rVert}
\newcommand{\mpart}[2]{\frac{\partial#1}{\partial#2}}

\newcommand{\comment}[1]{}

\theoremstyle{definition}

\newtheorem{assumption}{Assumption}

\newcommand{\pinv}[1]{{#1}^{\dagger}}
\begin{document}
\title{CASCLIK: CasADi-Based Closed-Loop Inverse Kinematics}
\author{Mathias~Hauan~Arbo,~\IEEEmembership{Member,~IEEE},
  Esten~Ingar~Gr{\o}tli,~\IEEEmembership{Member,~IEEE} and
  Jan~Tommy~Gravdahl,~\IEEEmembership{Senior Member,~IEEE}%
\thanks{M. H. Arbo and J. T. Gravdahl are with Department of
    Engineering Cybernetics, NTNU, Norwegian University of Science and
    Technology.}%
\thanks{E. I. Gr{\o}tli is with Mathematics and Cybernetics, SINTEF
    DIGITAL, Trondheim, Norway}%
\thanks{The work reported in this paper was supported by the centre
    for research based innovation SFI Manufacturing in Norway. The
    work is partially funded by the Research Council of Norway under
    contract number 237900.}}
\maketitle
\begin{abstract}
  A Python module for rapid prototyping of constraint-based
  closed-loop inverse kinematics controllers is presented. The module
  allows for combining multiple tasks that are resolved with a
  quadratic, nonlinear, or model predictive optimization-based
  approach, or a set-based task-priority inverse kinematics
  approach. The optimization-based approaches are described in
  relation to the set-based task approach, and a novel
  multidimensional ``in tangent cone'' function is presented for
  set-based tasks. A ROS component is provided, and the controllers
  are tested with matching a pose using either transformation matrices
  or dual quaternions, trajectory tracking while remaining in a
  bounded workspace, maximizing manipulability during a tracking task,
  tracking an input marker's position, and force compliance.
\end{abstract}
\comment{Note to Practitioners:
\begin{abstract}
  In many industrial use cases, the manipulator is redundant with
  respect to the task it is to achieve. Placing a symmetrical cup on a
  table, or spray painting a part rarely requires the full six degrees
  of freedom to be defined to in order to achieve the task. This
  redundancy can be exploited to combine multiple tasks such as
  maximizing manipulability of the task or keeping the shoulder away
  from an obstacle, and task-based closed-loop inverse kinematic
  frameworks allow one to do so. As task-based frameworks can involve
  a significant developmental effort, rapid prototyping of task-based
  controllers can be used to evaluate whether the alternative
  technique can result in potential speed up of the task, faster
  execution, or increased functionality. 
\end{abstract}
}
\IEEEpeerreviewmaketitle

\section{Introduction}
\IEEEPARstart{R}{obots} perform tasks that involve interacting and
moving objects in Cartesian space by moving joints and motors. Finding
control setpoints in terms of the joint coordinates such that the
robot can achieve the desired task requires solving the inverse
kinematics problem. Inverse kinematics is fundamental to all robots,
and occurs in everything from humanoid service robots to 3D printers,
surgical robots to autonomous vehicles. In this article we present
CASCLIK, a Python module for rapid prototyping of closed-loop inverse
kinematics controllers for realizing multiple constraint-based tasks.

Closed-loop inverse kinematics involves defining a feedback controller
for achieving the desired task. In \cite{Sciavicco1986}, Sciavicco et
al. present a closed-loop inverse kinematic approach where joint speed
setpoints minimize the distance to a given end-effector pose.  The
distance errors have a guaranteed convergence characteristics. The
controller works by inverting the differential kinematics and defines
a continuous motion control of the robot.

The task function approach by Samson et al. \cite{Samson1991}
describes a robotic task as defined by an arbitrary output function
and a control objective. The output function is a mapping from the
joint states and time to an output space.  Samson formulates the task
such that the control objective is to bring the output function to
zero.  The task function approach generalizes to a large class of
tasks as the output function may go from any robot states to any
positions or orientations defined relative the robot or world frame.

The constraint-based task specification approach of De~Schutter et
al. \cite{DeSchutter2007} describes procedures for designing tasks
with complex sensor-based robot systems and geometric
uncertainties. Constraint-based task specification uses variables
termed \emph{feature variables} to describe position of geometric and
task related features that are useful for the task. A key aspect of
constraint-based task specification is to allow for feature variables.

A robot is redundant with respect to its task when it has more degrees
of freedom than there are dimensions in the output function of the
task. This allows one to utilize the free degrees of freedom to
achieve tasks simultaneously. A common approach to handling redundancy
involves inverting the differential kinematics using a
pseudo-inverse. The pseudo-inverse often introduces a null-space
within which additional tasks can be achieved. To the author's
knowledge, the earliest article combining multiple tasks in this
manner is by Hanafusa et al. \cite{Hanafusa1981} where a 7
degrees-of-freedom robot tracks a trajectory and avoids an
obstacle. This is achieved by placing the lower priority tracking task
in the null-space of the higher priority obstacle avoidance task.
Chiaverini et al. shows in \cite{Chiaverini1997} that multiple tasks
can be combined in a singularity robust way. Any framework that
supports closed-loop inverse kinematics using task specification
should allow for multiple tasks. The state of the art presents two
approaches to multiple tasks, strict prioritization with null-space
based approaches such as the set-based singularity robust
task-priority inverse kinematics framework \cite{Moe2016} and
optimization-based prioritization which lacks strict priority but
allows prioritization through the cost function in an optimization
problem \cite{Aertbelien2014}.

Calculating the Jacobians involved in closed-loop inverse kinematics
has been a complicated process requiring explicit knowledge of the
underlying representation used in the tasks. Modern algorithmic
differentiation systems such as CasADi \cite{Andersson2013b} simplify
this process, allowing us to generate compiled functions of
complicated Jacobians. CasADi uses a symbolic framework for performing
algorithmic differentiation on expression graphs to construct
Jacobians. CasADi provides methods for formulating linear, quadratic,
and nonlinear problems that can be solved with e.g. QPOASES
\cite{Ferreau2014} and IPOPT \cite{Wachter2006}. CASCLIK translates a
set of tasks to optimal problems of a form that CasADi can solve. This
allows the user to test constraint-based programming with any of the
available optimizers in CasADi with the different controller
formulations presented in this article. The purpose of CASCLIK is to
facilitate rapid prototyping of constraint-based control of robot
systems.

The architecture of CASCLIK is inspired by eTaSL/eTC
\cite{Aertbelien2014} by Aertbeli\"{e}n et al., which is a more mature
task specification language and controller.  A core principle of the
architecture of eTaSL/eTC is to separate the low-level robot
controller, numerical solver, and the task specification. Tasks are
robot-agnostic and transferrable to any robot system with known
forward kinematics. The power of constraint-based task specification
and control has allowed the creation of a system architecture capable
of exploiting CAD knowledge for assembly \cite{Arbo2018CASE}, for
which this work may present alternative controller formulations of
interest. Robot-agnostic task specification enables execution of the
same task with different robot platforms, which also allows for easier
delegation of tasks to the appropriate robots, and transferral of
skills from one robot system to another.

CASCLIK is a CasADi-based Python module for testing closed-loop
inverse kinematics controllers. The module focuses on being
cross-platform and defers to CasADi for the symbolic backend and
optimization. The purpose of this module is to explore alternative
controller and constraint formulations that utilize the same general
structure as eTaSL/eTC. It considers nonlinear and model predictive
formulations which are less real-time applicable, in an attempt to
investigate aspects that may later be implemented into more
industrially relevant frameworks. As it uses CasADi for optimization,
CASCLIK utilizes the development efforts of the CasADi community to
enable a variety of solvers.

The article is divided into six sections. The first section introduces
relevant concepts such as closed-loop inverse kinematics, task
function approach,  algorithmic differentiation, and presents modern
related research. The second section describes the theory involved in
CASCLIK. The third section gives a brief description of the
implementation. The fourth section gives example applications of
CASCLIK and preliminary studies. The fifth and sixth section is the
discussion and conclusion.

The main contributions of the article are:
\begin{itemize}
\item A nonlinear programming formulation of the constraint-based
  closed-loop inverse kinematics task controller,
\item a model predictive formulation of the constraint-based
  closed-loop inverse kinematics task controller,
\item a general implementation of the set-based singularity robust
  multiple task-priority inverse kinematics framework of
  \cite{Moe2016},
\item a novel multidimensional \emph{in tangent cone} function for the
  set-based singularity robust multiple task-priority inverse
  kinematics framework,
\end{itemize}

\subsection{Related Research}
When considering fundamental robotics problems such as inverse
kinematics, there are innumerable important references. To limit the
scope we focus on related modern frameworks.

Stack-of-Tasks \cite{Mansard2009,stack-of-tasks} is a C++ software
development kit for real-time motion control of redundant
robots. Tasks and robots are defined using \emph{dynamic graphs} that
allow for caching results in functions for fast evaluation. The system
allows for equality and set tasks by activating and deactivating
control of the set tasks. The framework allows for joint torque level
control of the robot. Stack-of-tasks also supports a hierarchical
quadratic programming formulation \cite{Escande2014}. It is
open-source, includes tools for integration with ROS, and is limited
to Unix platforms.

iTaSC \cite{Smits2008,itasc-url} is a software framework for
constraint-based task specification and execution. It presents a
modular design for task specification, scenegraph representation, and
solver. The software framework is a part of the OROCOS project, and
uses OROCOS RTT \cite{Bruyninckx2003,rtt-url} to control robots.

The previously mentioned eTaSL/eTC \cite{Aertbelien2014} is a
successor to iTaSC, and is a C++/LUA constraint-based task
specification and control framework. Expressions are formulated using
\emph{expressiongraphs} \cite{expressiongraphs-url}, a symbolic
framework that uses OROCOS KDL definitions \cite{kdl-url} for frames
and rotations. Arbitrary symbolic expressions are used in constraints
to form a task specification. The architecture of eTaSL/eTC is
modular, allowing one to define new controllers for a task
specification and new solvers if they have a C++ interface.  It
currently supports QPOASES and the hierarchical quadratic programming
solver of Stack-of-Tasks. eTaSL/eTC includes a Python interface for
rapid prototyping and an OROCOS RTT \cite{Bruyninckx2003,rtt-url}
component for real-time control of robots using OROCOS. eTaSL/eTC is
open-source and is currently limited to Linux platforms.

Other advanced constraint-based approaches include the task level
robot programming framework of Somani et al.\cite{Somani2016}, that
supports an optimization based solver, and an analytical solver
\cite{Somani2017}. The software focuses on semantic process
description and CAD level tasks and constraints
\cite{Perzylo2016}. The CAD level constraints have composition rules,
allowing for a reduction of the space of possible control
setpoints. The reduced space is used to formulate the analytical
solver. To the author's knowledge, the software is not open-source.

The set-based singularity robust multiple task-priority inverse
kinematics controller \cite{Moe2016} is a task controller that uses
the augmented null-space projection operator \cite{Antonelli2009} and
activation or deactivation of null-spaces to implement set tasks. This
controller forms the null-space approach in CASCLIK and this article
extends the approach with support for multidimensional set
constraints.

\section{Theory}
In this section we present the underlying theory used in CASCLIK. We
present the variables and output function involved, the available
control objectives one can define, the convexity of the constraints in
optimization based controllers, and their effect in the null-space
projection based controller. Then we present the four different
controllers available: the quadratic, nonlinear, and model predictive
optimization-based approaches, and the null-space projection
approach. The quadratic programming approach is based on eTaSL/eTC
\cite{Aertbelien2014} and the null-space approach is based on the
set-based singularity robust task-priority inverse kinematics
controller \cite{Moe2016}.

\subsection{Variables and Output Function}
CASCLIK currently supports four different variable types:
\begin{itemize}
\item $t$, time,
\item $\bm{q}(t)\in\mR^{n_q}$, robot variable (e.g. joint angles),
\item $\bm{x}(t)\in\mR^{n_x}$, virtual variable (e.g. path-timing),
\item $\bm{y}(t)\in\mR^{n_y}$, input variables (e.g. sensor values).
\end{itemize}
Time and robot variables are self-explanatory. Virtual variables are
similar to the feature variables of eTaSL/eTC or iTaSC, but the term
feature implies a relation to geometric aspects of the task. We
describe these as virtual variables as they are variables maintained
by the computer, and not necessarily linked to any features of the
objects involved. This is merely a semantic choice. Virtual variables
simplify task specification and are present in cases such as
path-following. Input variables are variables for which we have no
information about the derivative behavior.

The output function is a function:
\begin{equation}
  \label{eq:output-function}
  \bm{e}(t,\bm{q},\bm{x},\bm{y})\in\mR^{n_e}, 
\end{equation}
where $n_e\leq n_q+n_x$ and $t$, $\bm{x}$ and $\bm{y}$ are optional.
In CASCLIK we assume no knowledge of the underlying geometry involved
when evaluating the partial derivatives of the output function. This
differs from most other closed-loop inverse kinematics frameworks
where the representation is used when evaluating the derivative of
transformation matrices and orientations. This is a design choice to
make the library as general as possible and allows us to inspect the
behavior with different representations. This may require more from
the task programmer as the behavior of the robot may differ depending
on the formulation of the output function.

\begin{assumption}[Velocity Control]\label{assumption:velocity-control}
  The robot system is equipped with a sufficiently fast velocity
  controller giving $\dot{\bm{q}}(t)=\dot{\bm{q}}_{des}(t)$ where
  $\dot{\bm{q}}_{des}$ is the designed control setpoint. The velocity
  controller controls all robot state velocities.
\end{assumption}

Samson et al.~\cite{Samson1991} describe how the first industrial
robots had velocity-controlled electrical motors, leading to the joint
velocity becoming the \emph{``true control variable''} for robot
systems in the control literature. Assumption
\ref{assumption:velocity-control} stems from this time and has been a
common robotics assumption since.

\subsection{Constraints}
We use a formulation of robotic tasks similar to Samson et
al.\cite{Samson1991}: a task is defined by an output function and a
control objective. Samson et al. defines the control objective as a
regulation problem where a task is performed perfectly during
$[t_0,t_f]$ if
\begin{equation}
  \label{eq:regulation-problem}
  \bm{e}(t,\bm{q},\bm{x},\bm{y})=0
\end{equation}
for all $t\in[t_0,t_f]$. This is achieved by designing a controller
such that the output function converges to zero.

Similar to eTaSL, we refer to the control objective as a type of
constraint. CASCLIK specifies four types of constraints:
\begin{itemize}
\item equality constraints,
  \begin{equation}
    \label{eq:equality-constraint}
    \bm{e}(t,\bm{q},\bm{x},\bm{y}) = 0,
  \end{equation}
\item set constraints,
  \begin{equation}
    \label{eq:set-constraint}
    \bm{e}_{l}(t,\bm{q},\bm{x},\bm{y}) \leq \bm{e}(t,\bm{q},\bm{x},\bm{y}) \leq \bm{e}_{u}(t,\bm{q},\bm{x},\bm{y}),
  \end{equation}
\item velocity equality constraints,
  \begin{equation}
    \label{eq:velocity-equality-constraint}
    \dot{\bm{e}}(t,\bm{q},\bm{x},\bm{y}) = \dot{\bm{e}}_{d}(t,\bm{q},\bm{x},\bm{y}),
  \end{equation}
\item and velocity set constraints,
  \begin{equation}
    \label{eq:velocity-set-constraint}
    \dot{\bm{e}}_l(t,\bm{q},\bm{x},\bm{y})\leq\dot{\bm{e}}(t,\bm{q},\bm{x},\bm{y})\leq\dot{\bm{e}}_u(t,\bm{q},\bm{x},\bm{y}).
  \end{equation}
\end{itemize}
where subscript $l$ and $u$ refer to the lower and upper bounds, and
subscript $d$ refers to a desired derivative of the output function.
The control objectives of the tasks are achieved perfectly if the
equations hold during $t\in[t_0, t_f]$.

As the control objectives both include equality (converging to zero),
and set constraints (converging to or remaining in a set), and set
constraints can have different upper and lower bounds, we cannot use
the regulation problem formulation of Samson et al. We rely on
linearization of the time-derivative of the output functions to
achieve the control objectives.

\begin{assumption}[Linearization]\label{assumption:linearization}
  The partial derivatives $\mpart{\bm{e}}{t}, \mpart{\bm{e}}{\bm{q}}$
  and $ \mpart{\bm{e}}{\bm{x}}$ (commonly called the task Jacobian)
  can be considered constant with respect to the control
  duration. That is:
  \begin{align}
    \mpart{\bm{e}}{t}(\tau) + &\mpart{\bm{e}}{\bm{q}}(\tau)\dot{\bm{q}}(t_n) + \mpart{\bm{e}}{\bm{x}}(\tau)\dot{\bm{x}}(t_n) \approx\nonumber\\
    &\mpart{\bm{e}}{t}(t_n) + \mpart{\bm{e}}{\bm{q}}(t_n)\dot{\bm{q}}(t_n) + \mpart{\bm{e}}{\bm{x}}(t_n)\dot{\bm{x}}(t_n) \label{eq:linearization-assumption}
  \end{align}
  for $t_n\leq\tau\leq t_n+\Delta_t$ where $t_n$ is a sampling point
  and $\Delta_t$ is the duration of the control step.
\end{assumption}

The linearization assumption is often used in closed-loop inverse
kinematics frameworks without explicitly stating it as an
assumption. The linearization assumption does not always hold, and for
long control steps or rapidly moving trajectories a tracking error may
occur \cite{Arbo2018_syroco}.

Defining a controller for a set of tasks is finding
$(\dot{\bm{q}},\dot{\bm{x}})$ such that we achieve the tasks. For the
optimization-based controller approaches we do this by imposing
constraints on the optimization problem, and for the null-space
approach we do this by both null-space projection and inversion of the
differential kinematics.

\subsubsection{Equality Constraints}
Taking the time-derivative of the output function, we get:
\begin{equation}
  \label{eq:output-function-derivative}
  \dot{\bm{e}} = \mpart{\bm{e}}{t} + \mpart{\bm{e}}{\bm{q}}\dot{\bm{q}} + \mpart{\bm{e}}{\bm{x}}\dot{\bm{x}}
\end{equation}
where the best guess for the derivatives of $\bm{y}$ is zero. An
equality constraint forms a regulation problem, which is achieved by
ensuring that:
\begin{equation}
  \label{eq:exponential-convergence-equality-constraint}
  \dot{\bm{e}} = \mpart{\bm{e}}{t} + \mpart{\bm{e}}{\bm{q}}\dot{\bm{q}} + \mpart{\bm{e}}{\bm{x}}\dot{\bm{x}} = -\bm{K}(t,\bm{q},\bm{x},\bm{y})\bm{e}(t,\bm{q},\bm{x},\bm{y})
\end{equation}
for which exponential convergence to zero is guaranteed if
\eqref{eq:exponential-convergence-equality-constraint} is upheld and
$\bm{K}$ is positive definite. $\bm{K}$ is a user-defined function,
and its dependent variables will be omitted for brevity in the rest of
the paper.

Velocity equality constraints are included to allow for velocity
following, but do not guarantee convergence:
\begin{equation}
  \label{eq:following-velocity-equality-constraint}
  \dot{\bm{e}} = \mpart{\bm{e}}{t} + \mpart{\bm{e}}{\bm{q}}\dot{\bm{q}} + \mpart{\bm{e}}{\bm{x}}\dot{\bm{x}} = \dot{\bm{e}}_{d}(t,\bm{q},\bm{x},\bm{y}),
\end{equation}
where the right hand side is the desired constraint velocity. This is
to accommodate control situations for which the desired output
function derivative is easy to define, but its integral is
not. Because we rely on Assumption \ref{assumption:velocity-control}
and Assumption \ref{assumption:linearization}, the lack of convergence
of velocity constraints is not considered in this article.

By using the Moore-Penrose pseudo-inverse (superscript $\dagger$) we
get \eqref{eq:exponential-convergence-equality-constraint} and
\eqref{eq:following-velocity-equality-constraint} on a
form that fits with the null-space projection approach. For equality
constraints it becomes:
\begin{equation}
  \label{eq:equality-constraint-pinv}
  \begin{bmatrix}
    \dot{\bm{q}}\\
    \dot{\bm{x}}
  \end{bmatrix} = -\pinv{\left(
    \begin{bmatrix}
      \mpart{\bm{e}}{\bm{q}} & \mpart{\bm{e}}{\bm{x}}
    \end{bmatrix}
\right)}\left(\bm{K}\bm{e}+\mpart{\bm{e}}{t}\right).
\end{equation}
Similarly for the velocity equality constraint we have:
\begin{equation}
  \label{eq:velocity-equality-constraint-pinv}
  \begin{bmatrix}
    \dot{\bm{q}}\\
    \dot{\bm{x}}
  \end{bmatrix} = \pinv{\left(
    \begin{bmatrix}
      \mpart{\bm{e}}{\bm{q}} & \mpart{\bm{e}}{\bm{x}}
    \end{bmatrix}
  \right)}\left(\dot{\bm{e}}_{d} - \mpart{\bm{e}}{t}\right).
\end{equation}

\subsubsection{Set Constraints}
\begin{figure}
  \centering
  \includegraphics[width=0.75\columnwidth]{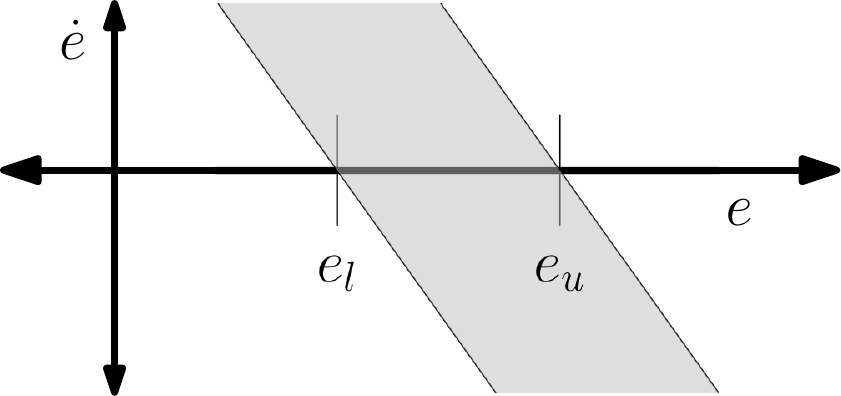}
  \caption{Visualization of set constraint convergence in one
    dimension.  $\dot{\bm{q}}$ and $\dot{\bm{x}}$ must be chosen such
    that $\dot{e}$ remains in the gray area. As this results in
    requiring $\dot{e}$ to be positive when $e < e_l$ and negative
    when $e > e_u$, we will converge to $e_l \leq e \leq e_u$. The
    slope of the lines are defined by $K$.}
  \label{fig:exponential-convergence-set-constraint}
\end{figure}

Set constraints are different in optimization approaches and the
null-space projection approach. In optimization based controllers we
enable exponential convergence to the set by defining the constraint
as:
\begin{align}
  \bm{K}(\bm{e}-\bm{e}_l)\leq \mpart{\bm{e}}{t}+&\mpart{\bm{e}}{\bm{q}}\dot{\bm{q}} + \mpart{\bm{e}}{\bm{x}}\dot{\bm{x}}\leq \bm{K}(\bm{e}-\bm{e}_u)  \label{eq:exponential-convergence-set-constraint}
\end{align}
where the gain is defined as previously.
Fig.\ref{fig:exponential-convergence-set-constraint} visualizes how
\eqref{eq:exponential-convergence-set-constraint} leads to
convergence. When approaching a limit from inside the constraint, the
maximum of $\dot{\bm{e}}$ will gradually be reduced, which causes an
exponential decay when approaching a constraint limit.

Similar to the velocity equality constraints, velocity set constraints
do not ensure convergence and are defined as:
\begin{equation}
  \label{eq:following-velocity-set-constraint}
  \dot{\bm{e}}_l(t,\bm{q},\bm{x},\bm{y}) \leq \mpart{\bm{e}}{t}+\mpart{\bm{e}}{\bm{q}}\dot{\bm{q}} + \mpart{\bm{e}}{\bm{x}}\dot{\bm{x}} \leq \dot{\bm{e}}_u(t,\bm{q},\bm{x},\bm{y})
\end{equation}
which can be visualized as horizontal lines in
Fig.\ref{fig:exponential-convergence-set-constraint}.

\begin{figure}
  \centering
  \includegraphics[width=0.75\columnwidth]{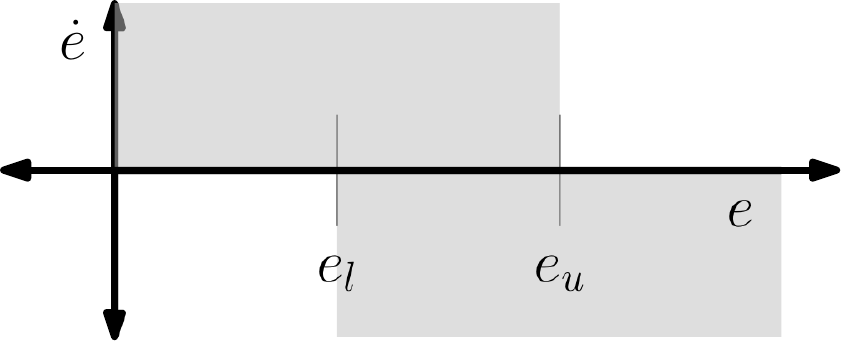}
  \caption{Visualization of when \emph{in tangent cone} evaluates to
    true for a one-dimensional output function. For any $(e,\dot{e})$
    not in the gray area, the set constraint becomes active and lower
    level tasks are projected into the null-space of the set
    constraint.}
  \label{fig:null-space-set-constraint}
\end{figure}

Set constraints in the null-space approach are handled using
null-space projection and the \emph{in tangent cone} function
(Algorithm 1 in \cite{Moe2016}). The method states that if the desired
$(\dot{\bm{q}},\dot{\bm{x}})$ ensures that the output function remains
in the set, by asking whether the $(e,\dot{e})$ pair is in the
extended tangent cone, then the set constraint is not active. If the
desired robot state velocity is not in the extended tangent cone then
the set constraint is active and lower priority tasks are modified by
the null-space projection operator of the active set constraint. The
choice of $(e,\dot{e})$ pairs that do not cause an activation of the
set constraint is visualized in
Fig.\ref{fig:null-space-set-constraint}. 

In \cite{Arbo2018_syroco} it was noted that formulating
multidimensional tracking tasks as one dimensional tasks of differing
priorities may lead to unexpected tracking errors. The \emph{in
  tangent cone} function of Moe et al. \cite{Moe2016} assumes
one-dimensional output functions. CASCLIK addresses this by
implementing a multidimensional version which allows for using
multidimensional output functions with set constraints. The algorithm
is given in Alg.\ref{alg:multidimensional-in-tangent-cone}. If at a
time we are at $\bm{e}$ then the vector $\bm{d}$ describes the normal
vector to the closest point that is in the set. This allows us to
identify when $\dot{\bm{e}}$ points inwards. The signs of
$\bm{e}-\bm{e}_l$ and $\bm{e}-\bm{e}_u$ are equal when the closest
point inside the set is on the corners of a set, allowing us to
identify the corners as special cases.
 \begin{algorithm}
  \caption{Multidimensional \emph{in tangent cone}}
  \label{alg:multidimensional-in-tangent-cone}
  \begin{algorithmic}[1]
    \renewcommand{\algorithmicrequire}{\textbf{Input:}}
    \REQUIRE $t,\bm{q},\bm{x},\bm{y},\dot{\bm{q}},\dot{\bm{x}}$\\
    \STATE $\bm{d}\leftarrow\textrm{sign}(\bm{e}-\bm{e}_l)+\textrm{sign}(\bm{e}-\bm{e}_u)$
    \STATE $\textrm{in\_crnr}\leftarrow\textrm{sign}(\bm{e}-\bm{e}_l)==\textrm{sign}(\bm{e}-\bm{e}_u)$
    \IF{$\bm{e}_l(t,\bm{q},\bm{x},\bm{y})\leq \bm{e}(t,\bm{q},\bm{x},\bm{y})\leq \bm{e}_u(t,\bm{q},\bm{x},\bm{y})$}
    \RETURN True
    \ELSIF{$\textrm{in\_crnr}$ \AND $|\bm{-d}\cdot\dot{\bm{e}}| < \mnorm{\bm{d}}\mnorm{\dot{\bm{e}}}\cos(45^{\circ}) $}
    \RETURN True
    \ELSIF{\NOT$\textrm{in\_crnr}$ \AND $\bm{d}\cdot\dot{\bm{e}}<0$}
    \RETURN True
    \ELSE
    \RETURN False
    \ENDIF
  \end{algorithmic}
\end{algorithm}

The multidimensional \emph{in tangent cone} function assumes that
corners can be approximated with a $45^{\circ}$ cone situated at the
corner. This is an approximation that may falsely report that we are
not in the extended tangent cone for $\textrm{dim}(\bm{e})>2$,
e.g. when the desired $\dot{\bm{e}}$ points along an edge of the set
constraint.

Velocity set constraints are not currently defined in the
task-priority inverse kinematics framework, and are therefore not
included in the null-space projection approach.

\subsubsection{Convexity of Desired Control Input Space}
For the optimization-based approaches, the task constraints form
constraints in the optimization problem. The derivative of the output
function is an affine function with respect to the desired control
input $\dot{\bm{q}}$ and $\dot{\bm{x}}$. For the different constraint
types, the space of desired control inputs are:
\begin{align}
  \mathcal{D}(\bm{e},eq) =&\left\{\dot{\bm{q}},\dot{\bm{x}}\bigg|\mpart{\bm{e}}{t}+\mpart{\bm{e}}{\bm{q}}\dot{\bm{q}}+\mpart{\bm{e}}{\bm{x}}\dot{\bm{x}} = -\bm{K}\bm{e}\right\}\label{eq:desired-equality-space}\\
  \mathcal{D}(\bm{e},vel.eq)=&\left\{\dot{\bm{q}},\dot{\bm{x}}\bigg|\mpart{\bm{e}}{t}+\mpart{\bm{e}}{\bm{q}}\dot{\bm{q}}+\mpart{\bm{e}}{\bm{x}}\dot{\bm{x}} = \dot{\bm{e}}_{d}\right\}\label{eq:desired-velocity-equality-space}\\
  \mathcal{D}(\bm{e},set)=&\nonumber\\
  \bigg\{ \dot{\bm{q}}, \dot{\bm{x}}\bigg|\mpart{\bm{e}}{t}+\mpart{\bm{e}}{\bm{q}}\dot{\bm{q}}&+\mpart{\bm{e}}{\bm{x}}\dot{\bm{x}} \in [\bm{K}(\bm{e}-\bm{e}_l), \bm{K}(\bm{e}-\bm{e}_u)]\bigg\}\label{eq:desired-set-space}\\
  \mathcal{D}(\bm{e},vel.set)=&\left\{\dot{\bm{q}},\dot{\bm{x}}\bigg|\mpart{\bm{e}}{t}+\mpart{\bm{e}}{\bm{q}}\dot{\bm{q}}+\mpart{\bm{e}}{\bm{x}}\dot{\bm{x}} \in [\dot{\bm{e}}_l, \dot{\bm{e}}_u]\right\}.\label{eq:desired-velocity-set-space}
\end{align}

At any particular time instance $t,\bm{q},\bm{x}$ and $\bm{y}$ are
constant, making $\dot{\bm{e}}$ an affine transformation with respect
to $\dot{\bm{q}}$ and $\dot{\bm{x}}$. From \cite{Boyd2006} we know
that the inverse image of an affine function on a convex set is
convex, which makes $\mathcal{D}$ convex. The set of possible choices
of $(\dot{\bm{q}},\dot{\bm{x}})$ with multiple tasks is
\begin{equation}
  \label{eq:possible-control-inputs}
  \bm{\mathcal{S}}(t,\bm{q},\bm{x},\bm{y}) = \bigcap_{i=1}^{n_{c}}\mathcal{D}(\bm{e}_i,c_i)(t,\bm{q},\bm{x},\bm{y}),
\end{equation}
where we have $n_c$ tasks, each with an output function $\bm{e}_i$ and
a control objective $c_i\in \{eq., vel.eq., set, vel.set\}$. As the
intersection of convex sets is convex, combining tasks maintains
convexity of the set of possible control variables. This convexity
hinges on the derivative of the output function being affine with
respect to the control variables. Thus the convexity argument does not
hold for the model predictive approach where $\bm{q}$ and $\bm{x}$ are
predicted variables for the predicted constraints.

Tasks are incompatible if $\mathcal{S}=\emptyset$ (e.g. first task is
to remain in a box, second task is to track a reference that leaves
the box). We can easily see that adding a slack variable term
$\bm{\epsilon}$ to $\dot{\bm{e}}$ reinstates the convexity with
respect to the variables
$(\dot{\bm{q}}, \dot{\bm{x}}, \bm{\epsilon})$ for the
non-predictive approaches.

\subsubsection{Null-Space Projection}
Given an output function $\bm{e}_i$, we define the null-space
projection operator of the task as:
\begin{equation}
  \label{eq:null-space}
  \bm{N}_{i}=\bm{I}_{n_q+n_x}-\pinv{\begin{bmatrix}
    \mpart{\bm{e}_i}{\bm{q}},\mpart{\bm{e}_i}{\bm{x}}
  \end{bmatrix}}\begin{bmatrix}
    \mpart{\bm{e}_i}{\bm{q}},\mpart{\bm{e}_i}{\bm{x}}
  \end{bmatrix}
\end{equation}
such that $\bm{N}_i\bm{v}=0$ if $\bm{v}$ is a vector that extends only
into the space of the task. For multiple tasks, the null-space of all
the tasks combined uses the augmented inverse-based projection of
\cite{Antonelli2009}:
\begin{equation}
  \label{eq:augmented-null-space}
  \bm{N}_{i,i+1,\dots}= \bm{I}_{n_q+n_x} - 
  \pinv{\begin{bmatrix}
        \mpart{\bm{e}_i}{\bm{q}},\mpart{\bm{e}_i}{\bm{x}}\\
        \mpart{\bm{e}_{i+1}}{\bm{q}},\mpart{\bm{e}_{i+1}}{\bm{x}}\\
        \vdots
  \end{bmatrix}}\begin{bmatrix}
        \mpart{\bm{e}_i}{\bm{q}},\mpart{\bm{e}_i}{\bm{x}}\\
        \mpart{\bm{e}_{i+1}}{\bm{q}},\mpart{\bm{e}_{i+1}}{\bm{x}}\\
        \vdots
  \end{bmatrix}
\end{equation}
With multidimensional set constraints the null-space should only
consider directions in which the output function violates the set
constraints. For each set constraint we define a diagonal activation
matrix $\bm{S}_i\in\mR^{n_{e_i}\times{n_{e_i}}}$ with diagonal
elements:
\begin{equation}
  \label{eq:boolean-vector}
  s_{i,j} = \left\{
    \begin{matrix}
      1, &  e_{i,j} < e_{l,i,j}\text{ or }e_{i,j}>e_{u,i,j}\\
      0, & \text{else}
    \end{matrix}
  \right.
\end{equation}
where subscript $i,j$ refers to $j$th element of the $i$th output
function, upper bound, or lower bound. With the activation matrix, the
augmented inverse-based projection becomes:
\begin{equation}
  \label{eq:augmented-null-space-multidim}
  \bm{N}_{i,i+1,\dots}= \bm{I}_{n_q+n_x} - 
  \pinv{\begin{bmatrix}
        \mpart{\bm{e}_i}{\bm{q}},\mpart{\bm{e}_i}{\bm{x}}\\
        \mpart{\bm{e}_{i+1}}{\bm{q}},\mpart{\bm{e}_{i+1}}{\bm{x}}\\
        \vdots
      \end{bmatrix}}
  \begin{bmatrix}
    \bm{J}_{A,i}\\
    \bm{J}_{A,i+1}\\
    \vdots
  \end{bmatrix}
\end{equation}
where
\begin{equation}
  \label{eq:multidim-jacobian}
  \bm{J}_{A,i} = \left\{
    \begin{matrix}
      [\mpart{\bm{e}_i}{\bm{q}}, \mpart{\bm{e}_i}{\bm{x}}], & \text{if equality constraint}\\
      \bm{S}_i[\mpart{\bm{e}_i}{\bm{q}}, \mpart{\bm{e}_i}{\bm{x}}], & \text{if set constraint}.
    \end{matrix}
  \right.
\end{equation}

\subsection{Quadratic Programming Approach}
The quadratic programming (QP) approach is a reactive control that
formulates a QP problem based on the current sensor information. At
time $t=t_k$, we know $\bm{q}(t_k)$,$\bm{x}(t_k)$, and $\bm{y}(t)$.
$\dot{\bm{q}}_k$ is a setpoint sent to the robot system and
$\dot{\bm{x}}_k$ is used to obtain $\bm{x}_{k+1}$. The optimization
problem is:
\begin{subequations}\label{eq:quadratic-programming-problem}
  \begin{align}
    \min_{\dot{\bm{q}}_k,\dot{\bm{x}}_k,\bm{\epsilon}}\quad &c\dot{\bm{q}}_k^T\bm{W}_{\dot{\bm{q}}}\dot{\bm{q}}_k + c\dot{\bm{x}}_k^T\bm{W}_{\dot{\bm{x}}}\dot{\bm{x}}_k + (1+c)\bm{\epsilon}^TW_{\bm{\epsilon}}\bm{\epsilon}\label{eq:quadratic-programming-cost}\\
    s.t.:\quad &\nonumber\\
               &(\dot{\bm{q}}_k,\dot{\bm{x}}_k)\in\mathcal{S}(t_k,\bm{q}(t_k),\bm{x}(t_k),\bm{y}(t_k),\bm{\epsilon})
  \end{align}
\end{subequations}
where $\mathcal{S}$ is the set of all $(\dot{\bm{q}}_k,\dot{\bm{x}}_k)$ such
that:
\begin{align}
  &\mpart{\bm{e}_i}{t} + \mpart{\bm{e}_i}{\bm{q}}\dot{\bm{q}}_k + \mpart{\bm{e}_i}{\bm{x}}\dot{\bm{x}}_k = -\bm{K}_i\bm{e}_i+\bm{\epsilon}_i\label{eq:qp-equality-constraints}\\
  &\bm{K}_j(\bm{e}_j-\bm{e}_{l,j})\leq\mpart{\bm{e}_j}{t}+\mpart{\bm{e}_j}{\bm{q}}\dot{\bm{q}}_k+\mpart{\bm{e}_j}{\bm{x}}\dot{\bm{x}}_k+\bm{\epsilon}_j\label{eq:qp-set-constraints-lower}\\` 
  &\mpart{\bm{e}_j}{t}+\mpart{\bm{e}_j}{\bm{q}}\dot{\bm{q}}_k+\mpart{\bm{e}_j}{\bm{x}}\dot{\bm{x}}_k+\bm{\epsilon}_j\leq \bm{K}_j(\bm{e}_j-\bm{e}_{u,j})\label{eq:qp-set-constraints-upper}\\
  &\mpart{\bm{e}_m}{t} + \mpart{\bm{e}_m}{\bm{q}}\dot{\bm{q}}_k + \mpart{\bm{e}_m}{\bm{x}}\dot{\bm{x}}_k = -\dot{\bm{e}}_{d,m}+\bm{\epsilon}_m\label{eq:qp-velocity-equality-constraints}\\
  &\dot{\bm{e}}_{l,n}\leq\mpart{\bm{e}_n}{t}+\mpart{\bm{e}_n}{\bm{q}}\dot{\bm{q}}_k+\mpart{\bm{e}_n}{\bm{x}}\dot{\bm{x}}_k+\bm{\epsilon}_n\leq \dot{\bm{e}}_{u,n}\label{eq:qp-velocity-set-constraints}
\end{align}
where $i\in[0,I]$ are all equality constraints, $j\in[0,J]$ are all
set constraints, $m\in[0,M]$ are all velocity equality constraints and
$n\in[0,N]$ are all the velocity set constraints.  $\bm{\epsilon}$
denotes a vector of all slack variables, and $c$ is a regularization
weight. The matrices $\bm{W}_{\dot{\bm{q}}},\bm{W}_{\dot{\bm{x}}}$ and
$\bm{W}_{\bm{\epsilon}}$ are positive definite matrices denoting the
weights on the robot velocity, virtual variable velocity, and the
slack variables respectively.

This formulation is based on the QP controller in eTaSL/eTC. The
gains, output functions, and partial derivatives of the output
functions are evaluated at time $t=t_k$ and assumed constant with
respect to the optimization problem. When the controller is started
for the first time, the virtual variables and the slack variables must
be initialized. This is done by solving the QP problem
\eqref{eq:quadratic-programming-problem} at time $t=t_0$ with
$\dot{\bm{q}}_0=0$.

The slack variable defines the behavior when constraints are
incompatible. This is a form of soft ``prioritization'' of the
constraints by avoiding the case where $\mathcal{S}=\emptyset$. With
the QP approach all objectives of the controller are formulated in
terms of constraints.

\subsection{Nonlinear Programming Approach}
The nonlinear programming (NLP) approach is a reactive control
approach that uses the same problem formulation as the QP approach,
but allows for more general cost expressions. The optimization problem
is:
\begin{subequations}\label{eq:nonlinear-programming-problem}
  \begin{align}
    \min_{\dot{\bm{q}}_k,\dot{\bm{x}}_k,\bm{\epsilon}}\quad & cf(t_k,\bm{q},\bm{x},\bm{y},\dot{\bm{q}}_k,\dot{\bm{x}}_k)+ (1+c)\bm{\epsilon}^TW_{\bm{\epsilon}}\bm{\epsilon}\label{eq:nonlinear-programming-cost}\\
    s.t.:\quad &\nonumber\\
               &(\dot{\bm{q}}_k,\dot{\bm{x}}_k)\in\mathcal{S}(t_k,\bm{q},\bm{x},\bm{y},\bm{\epsilon}_k)
  \end{align}
\end{subequations}
where the cost function $f$ is a user-defined function. If the cost
function is convex, then we have a convex optimization problem for
which efficient solvers exist. The cost function must depend upon
$\dot{\bm{q}}_k$, and $\dot{\bm{x}}_k$ if there are virtual variables,
the rest are optional.

As the QP and NLP approach are similar in their constraints and
regularization, any set of tasks implemented for the QP approach can
be implemented for the NLP approach. The NLP approach allows defining
more complex controllers by implementing objectives in terms of costs.

\subsection{Model Predictive Approach}
The QP and NLP approaches are reactive approaches where the current
states of the robot system are used to determine what the next control
input should be. A natural extension to such a system is the
introduction of model predictive control (MPC). The states are
predicted by relying on Assumption
\ref{assumption:velocity-control}. The MPC approach does not support
inputs $\bm{y}$ as there is no way of predicting what the input will
be.

The MPC approach problem is implemented using a multiple-shooting
strategy. We define the horizon length as $n_h$ steps of length
$\Delta_t$ and the optimization variable
$\bm{\chi}=\{\dot{\bm{q}}_k,\dot{\bm{x}}_k,\epsilon_k,\bm{q}_{k+1},\bm{x}_{k+1},\}_{k\in[0,n_h-1]}$. The
times are $t_k=t_0+\Delta_tk$. The control input duration is
$\Delta_t$. The problem is formulated as:
\begin{subequations}\label{eq:model-predictive-control-problem}
  \begin{align}
    \min_{\bm{\chi}}\quad & \Phi(\bm{\chi})\label{eq:model-predictive-control-cost}\\
    s.t.:\quad &\nonumber\\
    &\bm{q}_{0} = \bm{q}(t_0)\label{eq:mpc-initial-q}\\\
    &\bm{x}_{0} = \bm{x}(t_0)\label{eq:mpc-initial-x}\\
    &\bm{q}_{k+1}-(\bm{q}_k+\dot{\bm{q}}_k\Delta_t) = 0\label{eq:mpc-shooting-gap-q}\\
    &\bm{x}_{k+1}-(\bm{x}_k+\dot{\bm{x}}_k\Delta_t) = 0\label{eq:mpc-shooting-gap-x}\\
    &(\dot{\bm{q}}_k,\dot{\bm{x}}_k)\in\mathcal{S}(t_k,\bm{q}_k,\bm{x}_k)\label{eq:mpc-task-constraints}
  \end{align}
\end{subequations}
where $k\in[0,n_h]$ and 
\begin{align}
  \Phi(\bm{\chi}) = \sum_{k=0}^{n_h-1}&cf(t_k,\bm{q}_k,\bm{x}_k,\dot{\bm{q}}_k,\dot{\bm{x}}_k)+(1+c)\bm{\epsilon}_k^T\bm{W}_{\bm{\epsilon}}\bm{\epsilon}_k  \label{eq:model-predictive-control-cost-function}
\end{align}
is the cost from the NLP applied to each timestep along the prediction
horizon. \eqref{eq:mpc-initial-q}-\eqref{eq:mpc-initial-x} are lifting
conditions for the current
timestep. \eqref{eq:mpc-shooting-gap-q}-\eqref{eq:mpc-shooting-gap-x}
are the shooting gap constraints. $\mathcal{S}$ uses either the
shooting-gap variables for the predicted constraints or the numerical
value for the initial constraints. As mentioned, the convexity of the
task constraints is not guaranteed for the shooting-gap
variables. This is because the terms depend on predictions rather than
constants, and the derivative of the output function is not
necessarily affine.

The MPC approach is a bridge between closed-loop inverse kinematics
and motion planning. The MPC approach may utilize knowledge along its
prediction horizon to choose more appropriate control inputs. This
comes at the cost of computational complexity, and not guaranteeing
convexity of the constraints. With $n_q$ robot variables, $n_x$
virtual variables, $n_\epsilon$ slack variables, and the dimension of
the task constraints as $n_c$, the number of decision variables for
the QP and NLP approach is $n_q+n_x+n_{\epsilon}$ and there are $n_c$
constraint equations. For the MPC approach the number of decision
variables is $n_h(2n_q+2n_x+n_\epsilon)$ and the dimension of the
constraints is $n_h(2n_q+2n_x+n_c)$.

\subsection{Null-Space Projection Approach}
As previously stated, the null-space projection approach comes from
the singularity robust task-priority framework of Moe et
al.~\cite{Moe2016}. Tasks have a strict prioritization as lower
priority tasks are projected into the null-space of higher priority
tasks.

Given a priority sorted sequence $i\in [0, \dots, n_c-1]$ of
constraints, the desired control variables are:
\begin{equation}
  \label{eq:null-space-sequence}
    \begin{bmatrix}
    \dot{\bm{q}}\\
    \dot{\bm{x}}
  \end{bmatrix} = 
  \begin{bmatrix}
    \dot{\bm{q}}_{d,0}\\
    \dot{\bm{x}}_{d,0}
  \end{bmatrix} + \sum_{j=1}^{n_c-1}
  \bm{N}_{0,\dots,j}\begin{bmatrix}
    \dot{\bm{q}}_{d,j}\\
    \dot{\bm{x}}_{d,j}
  \end{bmatrix}
\end{equation}
where $[\dot{\bm{q}}_{d,0}^T,\dot{\bm{x}}_{d,0}^T]^T$ and
$[\dot{\bm{q}}_{d,j}^T,\dot{\bm{x}}_{d,j}^T]^T$ are defined by
\eqref{eq:equality-constraint-pinv} for equality constraints,
\eqref{eq:velocity-equality-constraint-pinv} for velocity equality
constraints, and $\bm{0}$ for set constraints. If a set constraint is
inactive, it does not contribute to the augmented null-space
projection of its lower priority tasks. If it is active, its task
Jacobian is used when formulating the null-space projector. The
null-space is formed using \eqref{eq:augmented-null-space} or
\eqref{eq:augmented-null-space-multidim} when using the
multidimensional \emph{in tangent cone} function. With $n_{set}$ set
based tasks, we have $2^{n_{set}}$ possible activation combinations of
the set constraints. These form $2^{n_{set}}$ possible modes of the
controller. In Table \ref{tab:null-space:activation-map} we see the
activation map of which map to activate based on whether the sets are
active, or inactive. This example has 3 set constraints, thus it has 8
modes. Check marks are active, dash marks are inactive set
constraints.
\begin{table}[h]
  \centering
  \caption{Activation map with 3 set constraints}
  \label{tab:null-space:activation-map}
  \begin{tabular}{c | c | c | c }
    \toprule
    Mode & Set 1& Set 2& Set 3\\
    \midrule
    1 & - & - & - \\
    2 & - & - & \checkmark\\
    3 & - & \checkmark & -\\
    4 & \checkmark & - & -\\
    5 & - & \checkmark & \checkmark\\
    6 & \checkmark & - & \checkmark\\
    7 & \checkmark & \checkmark & -\\
    8 & \checkmark & \checkmark & \checkmark
  \end{tabular}
\end{table}
At each timestep, we check each mode for whether the
$(\dot{\bm{q}},\dot{\bm{{x}}})$ it proposes is in the extended tangent
cone for all inactive set constraints. This is done by going down the
list of modes, activating set constraints until all other set
constraints are in their extended tangent cones. At worst we will
evaluate each mode and the times \emph{in tangent cone} is run is
$O(n_{set}\log_2(n_{set}))$\cite{sumof1s_oeis}.

The null-space projection approach gives hard limits on the
constraints. It ensures that we cannot chose control inputs that go
out of the sets. This differs from the optimization approaches where
the control variables are chosen such that the controller converges to
the sets. If external disturbances, numerical errors, or measurement
noises causes the null-space projection controller to end up outside
the edge of a set constraint, the controller will still attempt to
move along the null-space of the constraint, and not necessarily
converge into the set again. This means that one must start the
controller with the system inside the sets.

\section{Implementation}
In this section we briefly present the implementation of CASCLIK and
the support packages. This is to give insight into their purpose, and
important design choices.

\subsection{CasADi - Jacobian Damping}
As CasADi is a symbolic framework, it performs pseudo-inverse by
assuming that the item to be inverted has full rank. If $\bm{M}$ is
the matrix to be inverted CasADi solves the linear problem:
\begin{equation}
  \label{eq:casadi-pinv}
  \bm{M}\bm{M}^T\bm{x} = \bm{M}
\end{equation}
for $\bm{x}$ if $\bm{M}$ is wide, or by
\begin{equation}
  \label{eq:casadi-pinv-tall}
  \bm{M}^T\bm{M}\bm{x} = \bm{M}
\end{equation}
if $\bm{M}$ is tall.  We employ Jacobian damping \cite{Colome2015} to
give the Jacobian full rank as the activation matrix $\bm{S}_i$ in the
multidimensional set constraint may lead to zero rows in $\bm{J}_A$,
or ill-defined tasks may lead to zero rows. This modifies
\eqref{eq:casadi-pinv} to solve:
\begin{equation}
  \label{eq:casclik-pinv}
  (\bm{M}\bm{M}^T+\lambda\bm{I})\bm{x} = \bm{M}
\end{equation}
for wide matrices, where $\lambda$ is the damping factor, and similar
for tall matrices. The default damping factor is set to $10^{-7}$ and
is an option in the null-space approach controller.

\subsection{CASCLIK}
CASCLIK is a Python module that only depends on CasADi. This is to
have an operating system independent, robot middleware independent
software solution. The module is compatible with both Python 2.7 and
Python 3. CASCLIK is available on GitHub under the MIT
license \cite{mahaarbo-url}. The overall architecture is inspired by
eTaSL/eTC. The core module contains classes for constraints, skill
specification, and controllers.

The output function of a constraint is an arbitrary CasADi
expression. The gain, target derivative, and upper or lower limits can
be added depending on what type of control objective is
involved. Priority is added by specifying the constraint as soft or
hard for the optimization-based approaches, or as a numerical value
for the null-space based approach.

A collection of constraints is a skill. As the user is free to define
both what the time, robot, virtual, and input variables are called
when formulating the constraints, the user must provide the symbol for
each of the relevant variables to the skill specification as well as a
label and a list of constraints. The skill sorts the constraints
according to their numerical priority (relevant for the null-space
projection approach), and keeps track of whether there are slack
variables or virtual variables involved in the skill.

Controllers take a skill specification and other controller-dependent
parameters as well as an option dictionary. The
\emph{ReactiveQPController} (QP) takes a list of weights for the robot
or virtual variables. The \emph{ReactiveNLPController} (NLP) takes a
cost expression. The \emph{ModelPredictiveController} (MPC) takes a
cost expression as well as the horizon length, and a timestep
length. The null-space based \emph{PseudoInverseController} (PINV) has
no optional input.

The controllers are compiled using CasADi's just-in-time compilation
of solvers and functions. For problems containing a large number of
sets, PseudoInverseController has generally the longest compilation
time as there are $2^{n_{set}}$ separate modes to compile.

\subsection{Other modules}
Two additional modules were created to simplify
prototyping. \emph{urdf2casadi} is a Python module for generating
CasADi expressions for the forward kinematics of robots. It uses
either URDF files, which are common in ROS, or Denavit-Hartenberg
parameters, common in industry, for creating forward kinematics
reprented by a transformation matrix or a dual
quaternion. \emph{casclik\_basics} provides classes for interfacing
with robots that maintain the virtual variables and subscribes to
joint and sensor topics. Its \emph{DefaultRobotInterface} publishes
joint position commands, and its \emph{URModernInterface} is
specifically intended for use with the \emph{ur\_modern\_driver}
\cite{Andersen2015} and publishes joint speed
commands. \emph{casclik\_basics} is available on GitHub under the MIT
License \cite{mahaarbo-url}.

\section{Examples}
The tests were performed on a computer with an Intel Xeon CPU E5-1650
v3 running Ubuntu 16.04 with ROS Kinetic Kame. In all the experiments
we use QPOASES for the QP controller, IPOPT for the NLP and MPC
controller, and Jacobian damping for the PINV controller (null-space
approach).

\subsection{Representation - Matrix or Quaternion}
In this example we are controlling a UR5 robot using either dual
quaternions or transformation matrices for frame representation. The
example uses the UR5 URDF with \emph{urdf2casadi} to determine forward
kinematics. The robot is simulated at joint velocity level with Euler
discretization as we rely on Assumption
\ref{assumption:velocity-control}.

Transformation matrices $\bm{T}\in\mathcal{T}\subset\mR^{4\times4}$
are composed of a rotation matrix
$\bm{R}\in\mathcal{R}\subset\mR^{3\times 3}$, and a displacement
vector $\bm{p}\in\mR^3$. Dual quaternions $\breve{Q}$ are composed of
\begin{equation}
  \label{eq:dual-quaternion}
  \breve{Q} = Q_R + \varepsilon Q_p
\end{equation}
where $Q_R\in\mathcal{Q}_{unit}$ is a unit quaternion for rotation and
$Q_p\in\mathcal{Q}$ a quaternion for displacement. $\varepsilon$ is
the dual unit which satisfies $\varepsilon\varepsilon=0$. Dual
quaternions can be represented by a vector such that
$\bm{\breve{Q}}\in\mR^8$, and in vector form the quaternion product of
two dual quaternions $\breve{Q}_c=\breve{Q}_a\otimes\breve{Q}_b$ can
be defined as:
\begin{equation}
  \label{eq:dual-quaternion-hamilton-operator}
  \bm{\breve{Q}}_c = \bar{\bm{H}}(\bm{\breve{Q}}_b)\bm{\breve{Q}}_a = \bm{H}(\bm{\breve{Q}}_a)\bm{\breve{Q}}_b
\end{equation}
where $\bar{\bm{H}},\bm{H}\in\mR^{8\times8}$ are matrices referred to as
the minus and plus Hamilton operator \cite{Figueredo2013}. 

The UR5 has $\bm{q}\in\mR^6$ joint angles forming the robot
variables, and forward kinematics described by $\bm{R}(\bm{q})$ for the rotation matrix,
$\bm{p}(\bm{q})$ for the displacement, and $\bm{\breve{Q}}(\bm{q})$
for the dual quaternion.

The task is for the end-effector frame to match a desired frame. The
desired frame is described by $(\bm{R}_d,\bm{p}_d)$ with transformation
matrices, and $\bm{\breve{Q}}_d$ with dual quaternions. Using rotation
and displacement we can define this as the task:
\begin{equation}
  \label{eq:frame-matching-transformation-matrix}
  \bm{e}_{T}(\bm{q}) =
  \begin{bmatrix}
    \bm{p}(\bm{q}) - \bm{p}_d\\
    \mnorm{\bm{R}_d^T\bm{R}(\bm{q}) - \bm{I}}_F
  \end{bmatrix}
\end{equation}
where the first line ensures convergence of position and the second
ensures convergence of the rotation. The second line uses the
orientation metric of Larochelle et al. \cite{Larochelle2007} using
the Frobenius norm.

For dual quaternions, we employ the strategy of Figueredo et
al. \cite{Figueredo2013}:
\begin{equation}
  \label{eq:frame-matching-dual-quaternion}
  \bm{e}_Q(\bm{q}) = \bar{\bm{H}}(\bm{\breve{Q}}_d)\bm{C}(\bm{\breve{Q}}_d-\bm{\breve{Q}}(\bm{q}))
\end{equation}
where $\bm{C}=\text{diag}(-1,-1,-1,1,-1,-1,-1,1)$ is the conjugate
operator for dual quaternions in vector form. As $\bm{Q}_d$ is
constant, we see that \eqref{eq:frame-matching-dual-quaternion}
becomes linear with respect to $\bm{\breve{Q}}(\bm{q})$.

The joints have hard set constraints such that $q_i\in[-2\pi,2\pi]$,
and $\dot{q}_i\in[-\pi/5,\pi/5]$. As the null-space based controller
does not support velocity set constraints, the applied $\dot{\bm{q}}$
is also saturated by the max speed. The example is simulated with a
desired frame at $\bm{p}_{des}=[0.5, 0, 0.5]^T$, with a roll of
$5^{\circ}$. The control duration is 8 ms, and corresponds to 125
Hz. The MPC approach has a prediction horizon of 10 control steps. All
cost functions are the same as for the QP approach.

\begin{figure}[h]
  \centering
  \begin{subfigure}[t]{\columnwidth}
    \centering
    \includegraphics[width=\columnwidth]{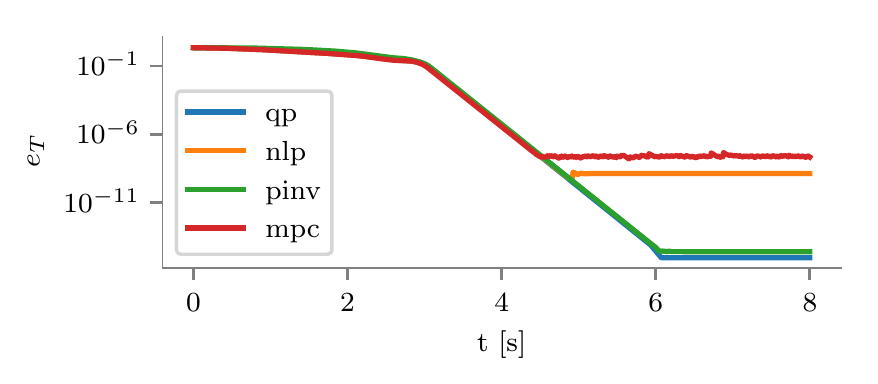}
    \caption{$\bm{e}_T$}
  \end{subfigure}
  \begin{subfigure}[b]{\columnwidth}
    \centering
    \includegraphics[width=\columnwidth]{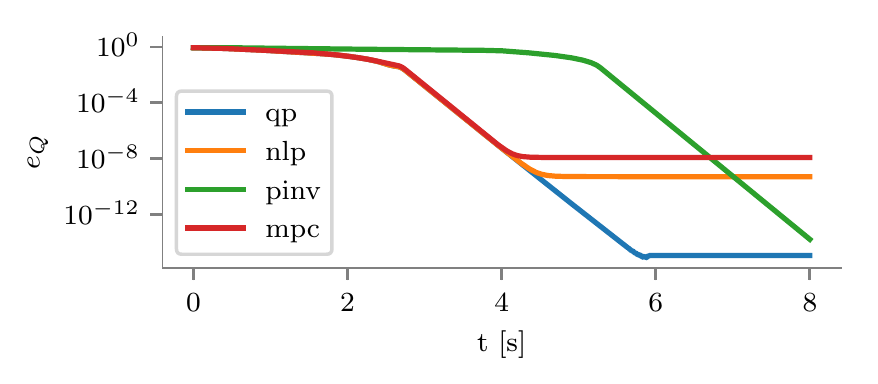}
    \caption{$\bm{e}_Q$}
  \end{subfigure}
  \caption{Euclidean norm of the output functions
    \eqref{eq:frame-matching-transformation-matrix} and
    \eqref{eq:frame-matching-dual-quaternion}.}
  \label{fig:examples:representation:errors}
\end{figure}

In Fig.\ref{fig:examples:representation:errors} we see the Euclidean
norm of the two representations for each of the controller
classes. The null-space approach has a greater error while moving
closer to the point as it does not account for the speed
saturation. It also struggles more with the dual quaternion
formulation and takes a more circuitous route. The different
controllers have different limits before numerical issues arise and
these may be optimizer settings dependent.

In Tab.\ref{tab:examples:representation:runtimes} the initial and
average runtimes are given for the different controllers during the
simulations. The null-space approach is denoted by PINV.

\begin{table}[h]
  \centering
  \caption{Controller runtimes for Representation Example}
  \label{tab:examples:representation:runtimes}
  \begin{tabular}{c c c c c}
    \toprule
    & PINV & QP & NLP & MPC\\ 
    \midrule
    Initial ($\bm{e}_Q$)& 0.11 ms& 0.95 ms& 4.23 ms& 26.80 ms\\
    Average ($\bm{e}_Q$)& 0.04 ms& 0.27 ms& 2.74 ms& 20.01 ms\\
    Initial ($\bm{e}_T$)& 0.09 ms& 1.44 ms& 4.37 ms& 16.90 ms\\
    Average ($\bm{e}_T$)& 0.04 ms& 0.26 ms& 2.81 ms& 147.08 ms
  \end{tabular}
\end{table}
The NLP approach uses approximately half the control duration, and MPC
approach generally uses an order of magnitude longer. This means that
the controllers would not be applicable to this control situation with
the default settings. The QP approach and null-space approach have
applicable timings with the QP approach being an order of magnitude
slower than the null-space approach. The average runtime of the MPC
formulation with transformation matrix formulation is much higher as a
result of using the Frobenius norm and matrix operations. This is
likely caused by the prediction constraints becoming more difficult to
apply.

\subsection{Set Constraints - Bounded Workspace}
This is a recreation of Example 2 from \cite{Moe2016} where a UR5
tracks a Cartesian trajectory while not escaping a box defined
workspace. The forward kinematics are defined by the
Denavit-Hartenberg parameters in \cite{Moe2016} using
\emph{urdf2casadi}.
\begin{figure}[h]
  \centering
  \includegraphics[width=0.7\columnwidth]{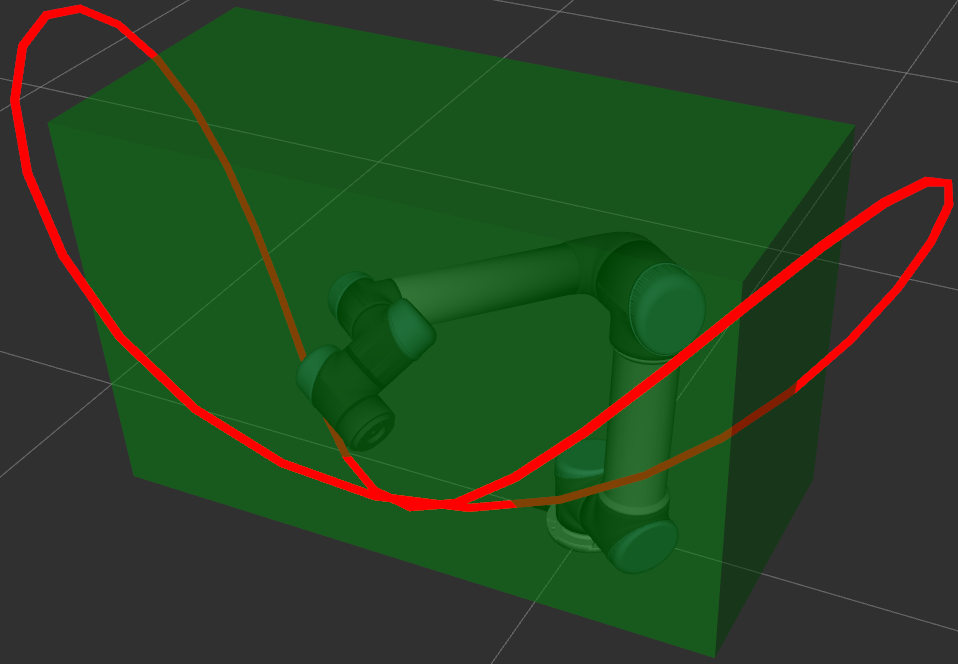}
  \caption{RViz visualization of the bounded workspace example. }
  \label{fig:examples:set-constraints:rviz}
\end{figure}

The output function is defined by
\begin{equation}
  \label{eq:examples:set-constraints:trajectory}
  \bm{e}(t,\bm{q}) = \bm{p}(\bm{q}) - \bm{p}_{des}(t)
\end{equation}
where $\bm{p}(\bm{q})$ is the forward kinematics to the origin of the
end-effector, and
\begin{equation}
  \bm{p}_{des}(t) = 
  \begin{bmatrix}
    0.5\sin^2(0.1t) + 0.2\\
    0.5\cos(0.1t) +0.25\sin(0.1t)\\
    0.5\sin(0.1t)\cos(0.1t)+0.7
  \end{bmatrix}.
\end{equation}

The set constraint is defined by
$\bm{p}(\bm{q})\in[\bm{p}_l,\bm{p}_u]$ with $\bm{p}_l=[0.1,-0.5,0.3]$
and $\bm{p}_l=[0.5,0.4,0.85]$. Examples of the code used to define the
equality and set constraints can be seen in
Listing.\ref{lst:examples:set-constraints:equality-code} and
Listing.\ref{lst:examples:set-constraints:set-code}, where
\texttt{p(q)} and \texttt{p\_d(t)} are CasADi functions for
end-effector position and desired position, and \texttt{t} and
\texttt{q} are MX symbols for time and robot variable. For the
null-space projection approach the set constraint can either be
formulated using the experimental multidimensional set constraint, or
as three separate constraints for $x$, $y$, and $z$ as in
\cite{Moe2016}. The equality constraint has a lower priority (3rd)
such that it can work with either formulation.

\begin{lstlisting}[caption=Equality Constraint Example,label={lst:examples:set-constraints:equality-code},float]
track_cnstr = EqualityConstraint(
    label="tracking_constraint",
    expression=p(q) - p_d(t),
    gain=1.0,
    constraint_type="soft",
    priority=3
)
\end{lstlisting}
\begin{lstlisting}[caption=Set Constraint Example,label={lst:examples:set-constraints:set-code},float]
box_cnstr = SetConstraint(
    label="box_constraint",
    expression=p(q),
    set_min = np.array([0.1, -0.5, 0.3]),
    set_max = np.array([0.5, 0.4, 0.85]),
    gain = 100.0,
    constraint_type="hard",
    priority=1
)
\end{lstlisting}

From Fig.\ref{fig:exponential-convergence-set-constraint} we know that
the approach speed to the upper or lower bound on a set constraint are
determined by the gain in the optimization approaches. This can be
seen as an exponential decay in the tracking task as we approach the
set limits. From Fig.\ref{fig:null-space-set-constraint} we see that
as the gain approaches $\infty$, we will have the same sharp change
when approaching a set limit as the null-space approach exhibits. In
this example the set gain is 100 for the QP and NLP. The MPC has gains
of 1 as large set gains may lead to more difficult predicted
constraints.

\begin{figure}[h]
  \centering
  \begin{subfigure}[t]{\columnwidth}
    \centering
    \includegraphics[width=\columnwidth]{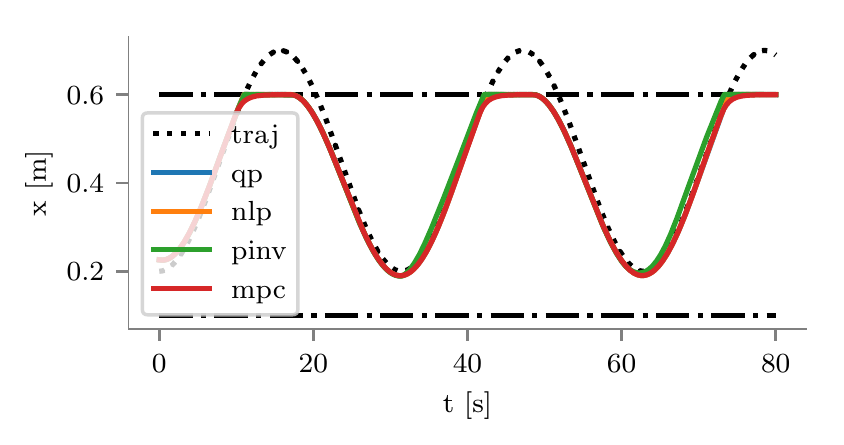}
    \caption{$x$}
  \end{subfigure}
  \begin{subfigure}[t]{\columnwidth}
    \centering
    \includegraphics[width=\columnwidth]{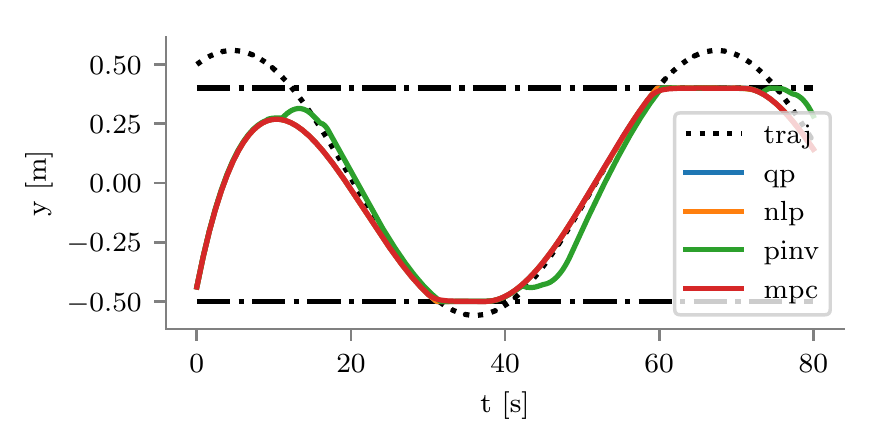}
    \caption{$y$}
  \end{subfigure}
  \begin{subfigure}[t]{\columnwidth}
    \centering
    \includegraphics[width=\columnwidth]{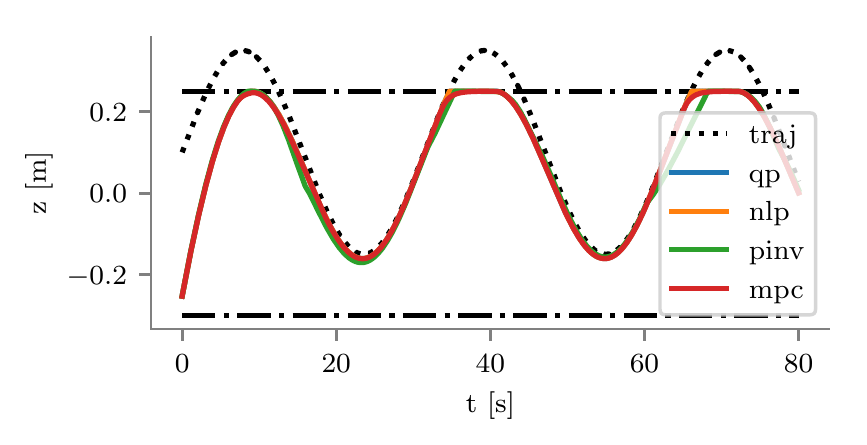}
  \end{subfigure}
  \caption{Position of the end-effector for $x$, $y$, $z$ controlled
    separately. $z$ omitted for brevity.}
  \label{fig:examples:set-constraints:singular-position}
\end{figure}

\begin{figure}[h]
  \centering
  \begin{subfigure}[t]{\columnwidth}
    \centering
    \includegraphics[width=\columnwidth]{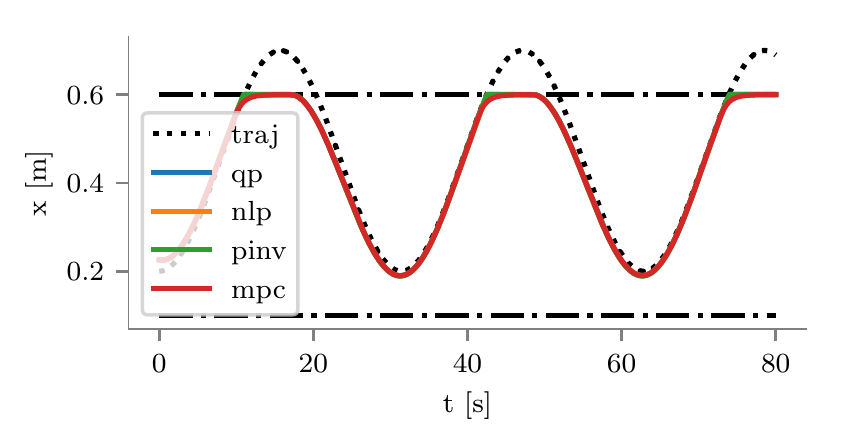}
    \caption{$x$}
  \end{subfigure}
  \begin{subfigure}[t]{\columnwidth}
    \centering
    \includegraphics[width=\columnwidth]{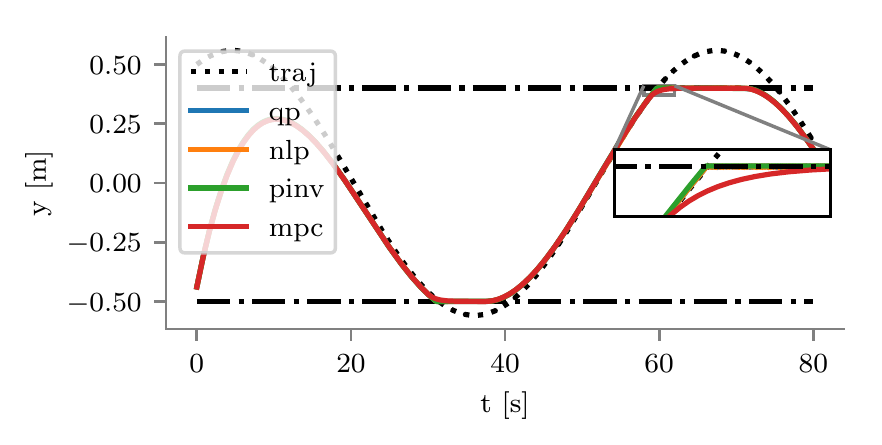}
    \caption{$y$}
  \end{subfigure}
  \begin{subfigure}[t]{\columnwidth}
    \centering
    \includegraphics[width=\columnwidth]{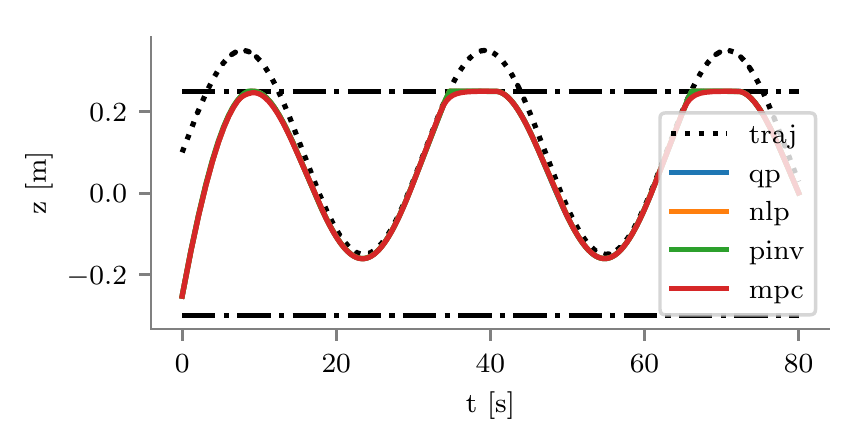}
    \caption{$z$}
  \end{subfigure}
  \caption{Position of the end-effector for $x$ , $y$, $z$ controlled
    using a multidimensional constraint. The MPC approach exhibits the
    exponential approach to the constraint limit.}
  \label{fig:examples:set-constraints:multidim-position}
\end{figure}

In Fig.\ref{fig:examples:set-constraints:singular-position} we see the
position of the end-effector for the different controllers when
handling $x$, $y$, and $z$ as separate constraints and in
Fig.\ref{fig:examples:set-constraints:multidim-position} we see the
position when handling them as a single multidimensional constraint.

\begin{figure}[h]
  \centering
  \includegraphics[width=\columnwidth]{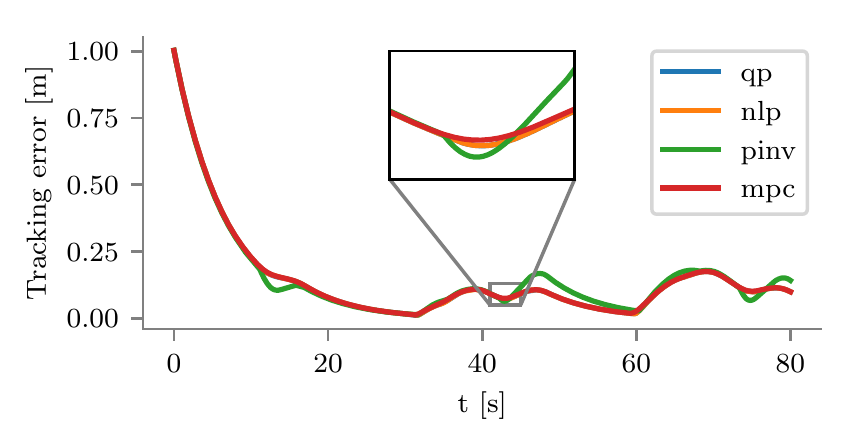}
  \caption{Tracking error for $x$, $y$, and $z$ as separate
    constraints. The error does not converge to zero as the desired
    trajectory goes out of the box.}
  \label{fig:examples:set-constraints:error-singular}
\end{figure}

\begin{figure}[h]
  \centering
  \includegraphics[width=\columnwidth]{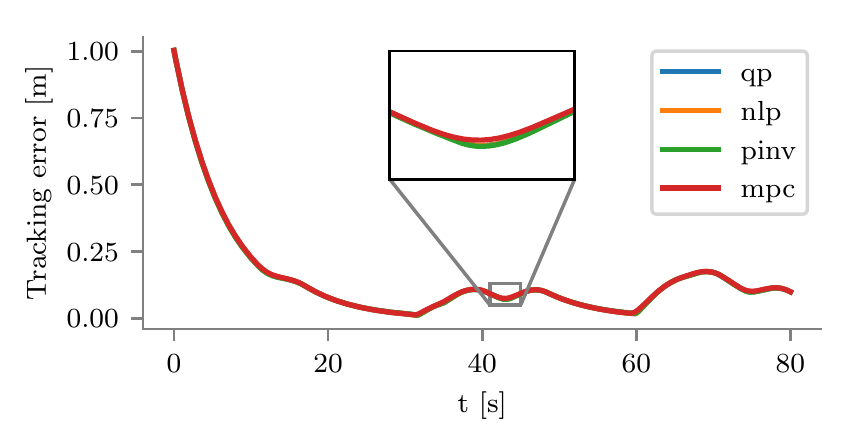}
  \caption{Tracking error for $x$, $y$ and $z$ as a multidimensional
    constraint. The error does not converge to zero as the desired
    trajectory goes outside the box. All are equal except MPC which
    deviates slightly.}
  \label{fig:examples:set-constraints:error-multidim}
\end{figure}

\begin{figure}[h]
  \centering
  \includegraphics[width=\columnwidth]{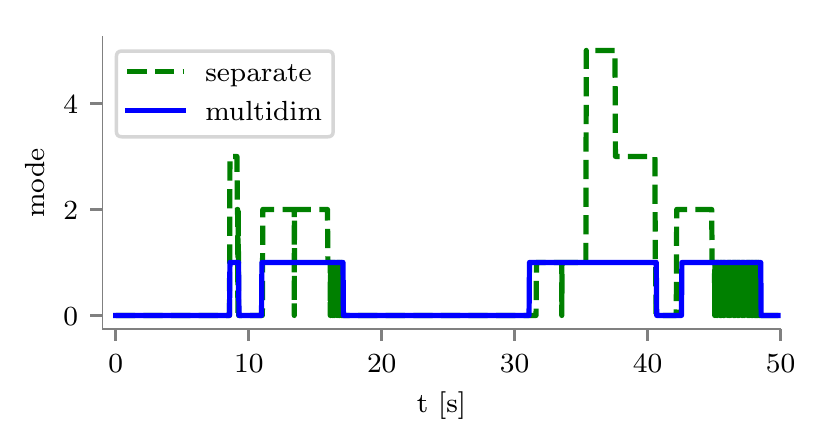}
  \caption{Excerpt of the mode the null-space based controller is in
    for both separate constraints and multidimensional
    constraints. Note the rapid switching at $t=15$s and $t=45$s when
    using separate constraints.}
  \label{fig:examples:set-constraints:modes}
\end{figure}

In Fig.\ref{fig:examples:set-constraints:error-singular} we see the
tracking error with the different controllers when handling $x$, $y$,
and $z$ constraints as separate constraints. In
Fig.\ref{fig:examples:set-constraints:error-multidim} we see the
tracking error with the different controllers when handling $x$, $y$,
and $z$ as a single multidimensional constraint. Similar to the
results in \cite{Arbo2018_syroco}, setting the constraints in a
priority ordered sequence causes unwanted behavior. The
multidimensional formulation gives a more correct interpretation of
the constraint. Also note that the multidimensional null-space
approach and the optimization approaches are similar. If more set
constraints are included, such as joint limits, the two methods will
differ again. In Fig.\ref{fig:examples:set-constraints:modes} we see
the mode the null-space based controller is in for both the separate
$x$, $y$, and $z$ formulation and the multidimensional
formulation. The ``noisy'' rapid switching of modes occurs due to
numerical issues with the linearization and the comparison between
current and set limits. Tuning either the control duration or the
comparison with some numerical lower limit can mitigate this effect.

Fig.\ref{fig:examples:set-constraints:rviz} shows a visualization of
the controller running with the DefaultRobotInterface and ROS.

\subsection{Nonlinear cost - Manipulability Index}
In this example a 7 degrees of freedom KUKA LWR IIWA 14 R820 arm is to
follow a circular trajectory in its workspace and maximize the
manipulability index of the task. The forward kinematics are defined
by the URDF and \emph{urdf2casadi}. The robot is simulated at joint
velocity level with Euler discretization. IIWA has $\bm{q}\in\mR^7$
where $\bm{q}\in[-\bm{q}_u,\bm{q}_u]$ with
\begin{equation}
  \label{eq:examples:nonlinear-cost:limits}
  \bm{q}_u^T=[170^{\circ}, 120^{\circ},170^{\circ}, 120^{\circ},170^{\circ}, 120^{\circ},175^{\circ}].
\end{equation}
The circular trajectory is defined by:
\begin{equation}
  \label{eq:examples:nonlinear-cost:circular-trajectory}
  \bm{p}_d(t) =
  \begin{bmatrix}
    0.1\cos(0.05t - 0.5\pi)+0.45\\
    0.1\sin(0.05t - 0.5\pi)+0.4\\
    0.3
  \end{bmatrix}.
\end{equation}
We define the manipulability index of the task as
\begin{equation}
  \label{eq:examples:nonlinear-cost:manipulability}
  m(\bm{q}) = \sqrt{\det\left(\mpart{\bm{p}}{\bm{q}}(\bm{q})\mpart{\bm{p}}{\bm{q}}(\bm{q})^T\right)}
\end{equation}
where $m$ is a measure of the area of the ellipsoid that the Jacobian
of the end-effector position forms. We want to both achieve the task
and to maximize our manipulability. Maximizing the manipulability can
be beneficial in collision avoidance, as a high manipulability means
we have more options as to which direction we can move to avoid
collision. Maximizing the manipulability can be achieved by adding a
term to the nonlinear costs of the NLP and the MPC approach:
\begin{equation}
  \label{eq:examples:nonlinear-cost:manipulability-cost}
  m_c(\bm{q},\dot{\bm{q}}) = -\alpha m(\bm{q}+\Delta_t\dot{\bm{q}})^2
\end{equation}
where $\Delta_t$ is the control duration and $\alpha$ is set to
$500$. This essentially states that we attempt to maximize the
manipulability of the subsequent step. Optimization problems often
struggle with square roots, so we square the manipulability index
before using it in the cost. As the tracking constraint is of lower
priority than the joint limits, we must ensure that the tracking
constraint's slack weight is greater than $\alpha$. In
Listing \ref{lst:examples:nonlinear-cost:equality-code}, we see an
example of setting its slack weight to $2000$. The QP approach and
null-space approach does not support nonlinear costs and do not
maximize their manipulability. The MPC has a horizon length of 10
control steps.

\begin{lstlisting}[caption=Equality Constraint With Slack Weight,label={lst:examples:nonlinear-cost:equality-code},float]
track_cnstr = EqualityConstraint(
    label="tracking_constraint",
    expression=p(q) - p_d(t),
    gain=1.0,
    constraint_type="soft",
    slack_weight=2e3
)
\end{lstlisting}

\begin{figure}
  \centering
  \includegraphics[width=\columnwidth]{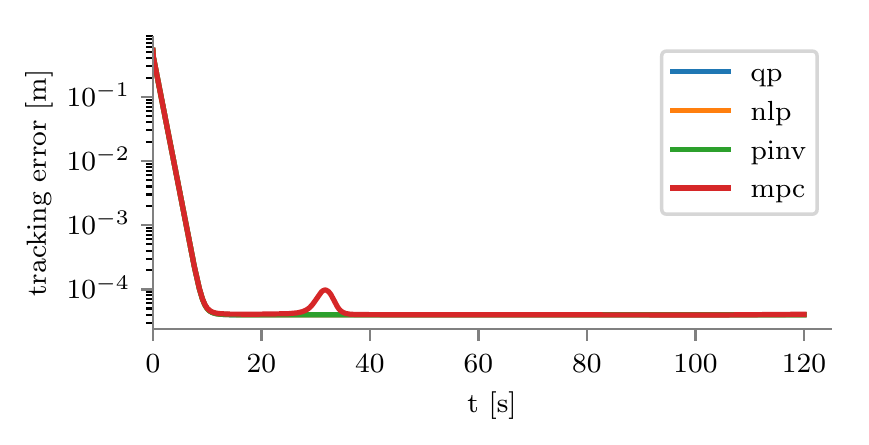}
  \caption{Tracking error when following the circular trajectory. The
    MPC approach deviates slightly at $t=30$s as it is reconfiguring
    to an orientation with higher manipulability.}
\label{fig:examples:nonlinear-cost:circ-error}
\end{figure}

In Fig.\ref{fig:examples:nonlinear-cost:circ-error} we see the
tracking error over time. The lower limit stems from the linearization
assumption, and one must either use a path-following approach, or use
lower control durations to overcome this. At $t=30$s, the MPC approach
is able to find a different configuration with higher manipulability
at the cost of a short duration of deviating from the trajectory. This
reconfiguration does not affect the final tracking error of the MPC.

\begin{figure}
  \centering
  \includegraphics[width=\columnwidth]{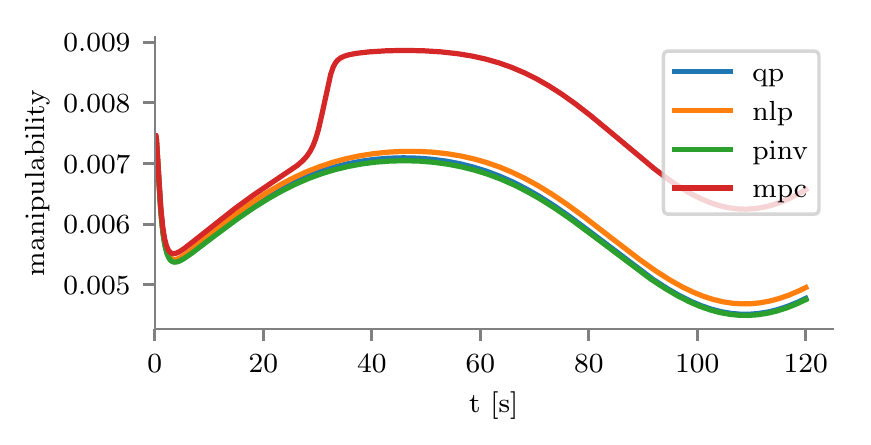}
  \caption{Manipulability index for different controllers over time
when tracking the circular trajectory. The MPC approach is
able to find a configuration with higher manipulability index at
$t=30s$.}
\label{fig:examples:nonlinear-cost:circ-manipulability}
\end{figure}

In Fig.\ref{fig:examples:nonlinear-cost:circ-manipulability} we see the
manipulability index $m$ over time. The NLP performs slightly better
than the QP approach, and the MPC performs best by far as it chooses
to deviate slightly from the trajectory to arrive at a configuration
with a higher manipulability. 

\begin{table}
  \centering
  \caption{Controller runtimes for Manipulability Example}
  \label{tab:examples:nonlinear-cost:runtimes}
  \begin{tabular}{c c c c c}
    \toprule
    & PINV & QP & NLP & MPC\\ 
    \midrule
    Initial& 0.11 ms& 0.48 ms& 4.94 ms& 54.72 ms\\
    Average& 0.07 ms& 0.19 ms& 3.02 ms& 49.52 ms
  \end{tabular}
\end{table}

In Tab.\ref{tab:examples:nonlinear-cost:runtimes} we see the initial
and average runtimes of the different controllers. The inclusion of
the manipulability cost has not made the controllers deviate from the
rule of an order of magnitude separation between the approaches. 

\subsection{Input - Tracking a marker in ROS}
\begin{figure}
  \centering
  \includegraphics[width=0.7\columnwidth]{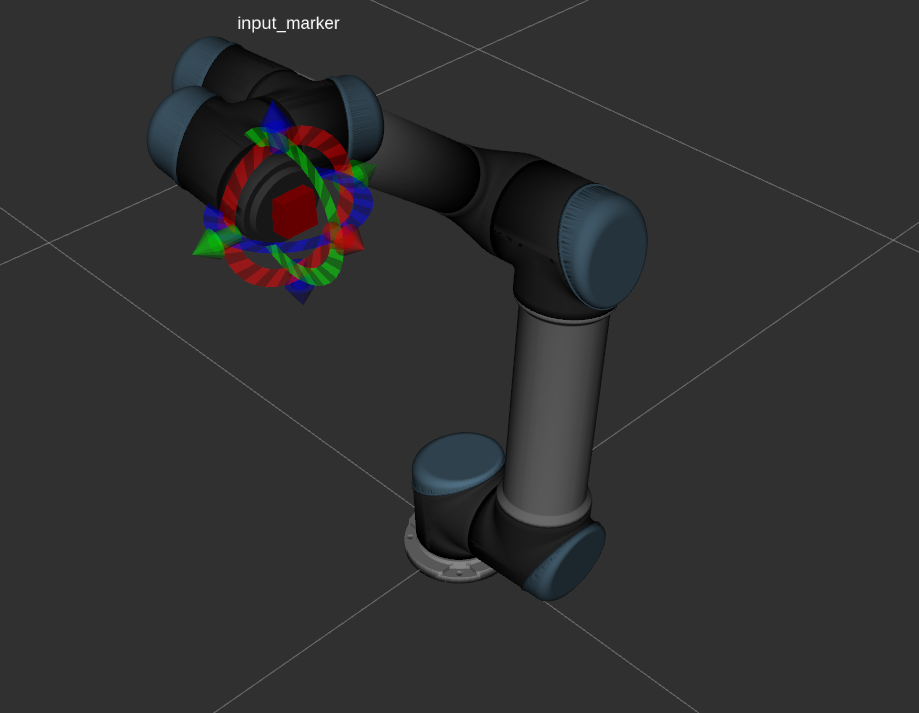}
  \caption{RViz visualization of tracking an input marker.}
  \label{fig:examples:input:rviz}
\end{figure}
In this example a UR5 robot tries to track user input. The
DefaultRobotInterface is used with ROS and an RViz interactive marker
to simulate an external input. The robot is simulated with Gazebo and
is controlled at 50 Hz. We use the QP approach in this example. The
maximum joint speed is $3$ rad/s.

\begin{figure}
  \centering
  \includegraphics[width=\columnwidth]{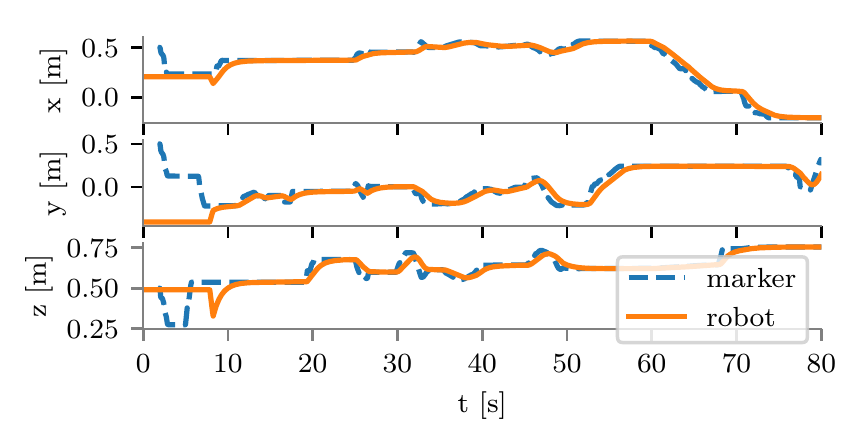}
  \caption{Position of the input marker and end-effector frame when
    tracking the input marker.}\label{fig:examples:input:position}
\end{figure}

In Fig.\ref{fig:examples:input:position} we see the position of the
marker and the end-effector frame. The end-effector has an exponential
convergence to the desired input marker position, but as it does not
consider the speed of the input marker, there is a tracking error when
following the input marker during a continuous motion. The
DefaultRobotInterface has a delay of 7.8s before it starts as it waits
for topics and compiles the controller.

\subsection{Velocity Equality Constraint - 6 DOF compliance}
\begin{figure}[h]
  \centering
  \includegraphics[width=0.7\columnwidth]{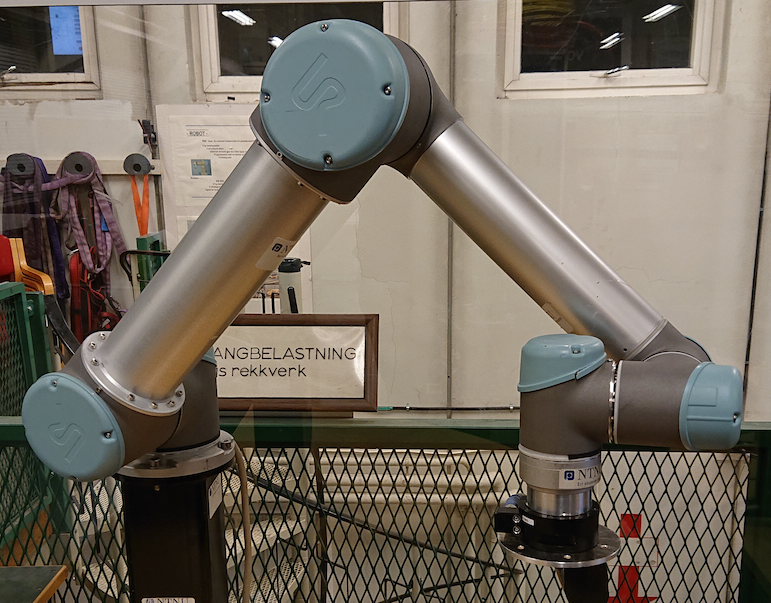}
  \caption{Experimental setup for 6 DOF compliance.}
  \label{fig:examples:velocityconstraint:experimental_setup}
\end{figure}
In this example the end-effector of a UR5 is to comply to forces and
torques acting on it. The example uses \emph{urdf2casadi} to determine
forward kinematics, and \emph{ur\_modern\_driver}\cite{Andersen2015}.
The end-effector has an ATI Mini45 force/torque sensor attached with a
mounting plate on it.  The robot is in an open workspace. To ensure
the robot does not crash with the table or is moved to undesired
regions, the end-effector is limited to a box. The box is defined by
the set constraint
\begin{equation}
  \label{eq:velocity-equality-constraint:box}
  \begin{bmatrix}
    -0.7\\
    -0.4\\
    -0.2
  \end{bmatrix}\leq
  \bm{p}(\bm{q})
  \leq
  \begin{bmatrix}
    -0.3\\
    0.5\\
    0.5
  \end{bmatrix}.
\end{equation}

Velocity resolved compliance can be achieved using damping control\cite{Siciliano2008} by
\begin{equation}
  \label{eq:examples:velocityconstraint:damping-control}
  \begin{bmatrix}
    \bm{v}(t)\\
    \bm{\omega}(t)
  \end{bmatrix}=
  \begin{bmatrix}
    K_{f}\bm{f}(t)\\
    K_{\tau}\bm{\tau}(t)
  \end{bmatrix}
\end{equation}
where $\bm{v}$ is the Cartesian velocity, $\bm{\omega}$ is the
rotational velocity, $\bm{f}$ are the linear forces, and $\bm{\tau}$
are the torques. All evaluated at the end-effector. $K_f$ and
$K_{\tau}$ are the damping constants for the forces and torques
respectively.

For linear forces $\bm{f}$ acting on the end-effector, and a position
$\bm{p}(t)$ of the end-effector, we desire:
\begin{equation}
  \label{eq:examples:velocityconstraint:force-comply}
  \dot{\bm{p}}(t) = K_f\bm{f}^w(t)= K_f\bm{R}(\bm{q})\bm{f}(t)
\end{equation}
where $\bm{f}^w$ are the forces acting on the end-effector represented
in the world coordinates.

We can relate the rotational velocity to the derivative of the
orientation quaternion by following \cite{Wang2012}, and arrive at
\begin{equation}
  \label{eq:examples:velocityconstraint:torque-comply}
  \dot{\bm{Q}}_r(t) = \frac{K_{\tau}}{2}\bar{\bm{H}}\left(
    \begin{bmatrix}
      \bm{\tau}(t)\\
      0
    \end{bmatrix}
\right)\bm{Q}_r(t)
\end{equation}
where $\bm{\tau}(t)$ are the torques acting on the end-effector in the
end-effector frame.

An example of the constraint is given in Listing
\ref{lst:examples:velocityconstraint:velocity-equality-code}.
\begin{lstlisting}[caption=Velocity Equality Constraint Example, label={lst:examples:velocityconstraint:velocity-equality-code},float]
comply_cnstr = VelocityEqualityConstraint(
    label="comply_constraint",
    expression=vertcat(p(q), Q_rot(q)),
    target=vertcat(Kf*p(q),
        0.5*Kt*mtimes(H(tau,0), Q_rot(q))),
    constraint_type="soft"
)
\end{lstlisting}

The controller is running at 125 Hz, and the force/torque sensor runs
at 250 Hz but only the most recent value is used. The damping factors
are $K_f=0.01$ and $K_t=0.1$. This experiment was run with the QP
controller. To inspect the behavior of the system we look at the right
hand side of \eqref{eq:examples:velocityconstraint:force-comply} and
\eqref{eq:examples:velocityconstraint:torque-comply} with the sensor
value for force, torque, and $\bm{q}$. We refer to this as
\emph{sensed speed}. The left hand side of
\eqref{eq:examples:velocityconstraint:force-comply} and
\eqref{eq:examples:velocityconstraint:torque-comply} as desired by the
CASCLIK controller or as reported by the robot, is referred to as the
\emph{controller speed} and the \emph{robot speed} respectively.

\begin{figure}
  \centering
  \includegraphics[width=\columnwidth]{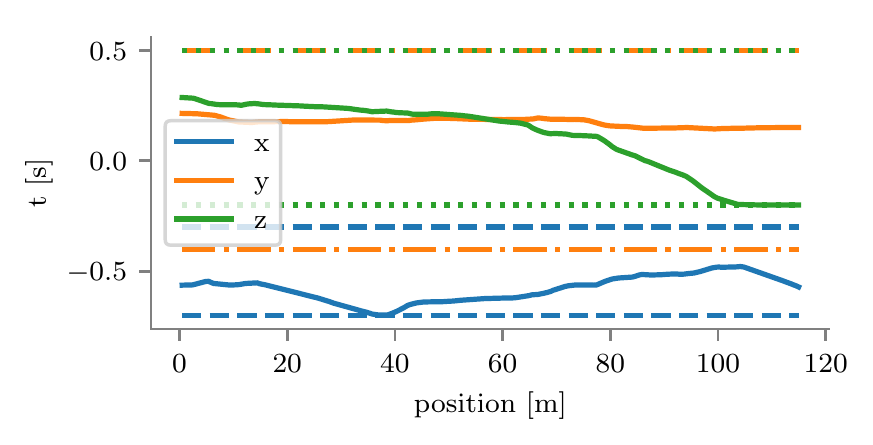}
  \caption{Position over time of the end-effector during the velocity
    equality experiment for compliance. The stippled and dotted lines
    denote the box constraint. From $t=36$s until $t=40$s we attempt
    to pull the end-effector out of the box constraint in $x$
    direction, but the set constraint does not allow it. At $t=80$s we
    let go of the end-effector, and it drifted to the bottom of the
    box due to sensor bias.}
  \label{fig:examples:velocityconstraint:position}
\end{figure}

\begin{figure}
  \centering
  \begin{subfigure}[t]{\columnwidth}
      \centering
      \includegraphics[width=\columnwidth]{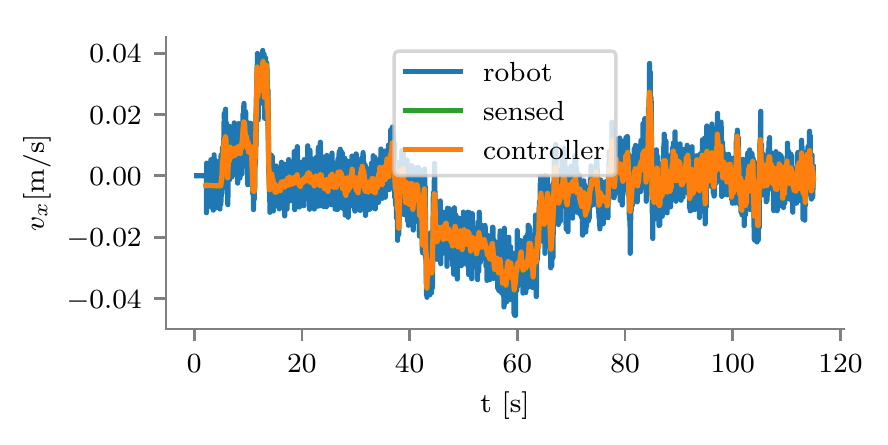}
      \caption{Complying with $f_x$}
    \end{subfigure}
  \begin{subfigure}[t]{\columnwidth}
    \centering
    \includegraphics[width=\columnwidth]{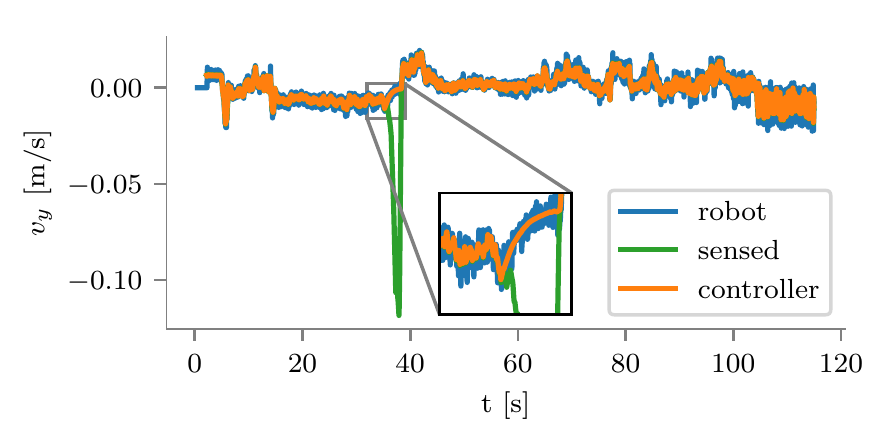}
    \caption{Complying with $f_y$}
  \end{subfigure}
  \begin{subfigure}[t]{\columnwidth}
    \centering
    \includegraphics[width=\columnwidth]{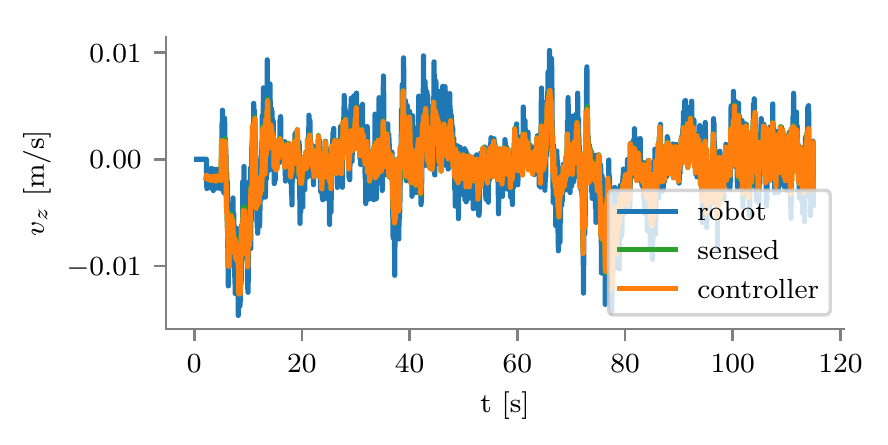}
    \caption{Complying with $f_z$}
  \end{subfigure}
  \caption{Sensor and controller Cartesian speeds (right and left hand
    side of \eqref{eq:examples:velocityconstraint:force-comply}), and
    robot Cartesian speeds of the end-effector. The zoomed inset at
    $t=36$s in subfigure (b) shows the exponential convergence to zero
    speed in $z$ direction as we try to pull the end-effector out of
    the box constraint. Otherwise, the sensed and controller speeds
    perfectly overlap as long as we are inside the box. }
  \label{fig:examples:velocityconstraint:forces}
\end{figure}

\begin{figure}
  \centering
  \begin{subfigure}[t]{\columnwidth}
      \centering
    \includegraphics[width=\columnwidth]{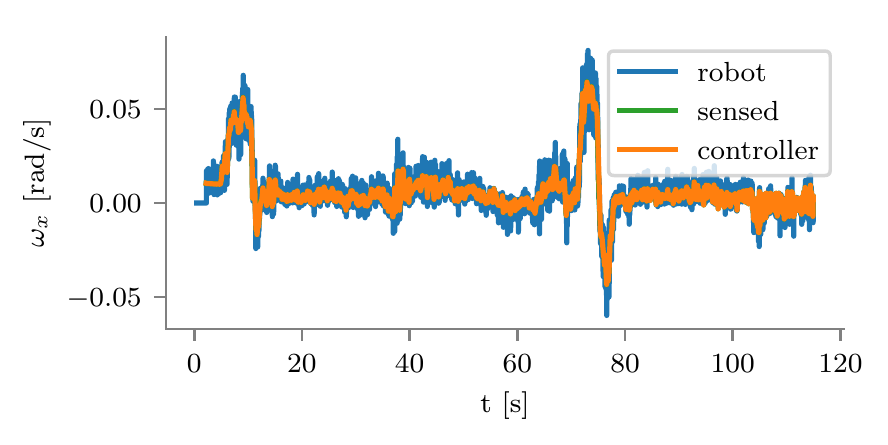}
    \caption{Complying with $\tau_x$}
  \end{subfigure}
\begin{subfigure}[t]{\columnwidth}
  \centering
    \includegraphics[width=\columnwidth]{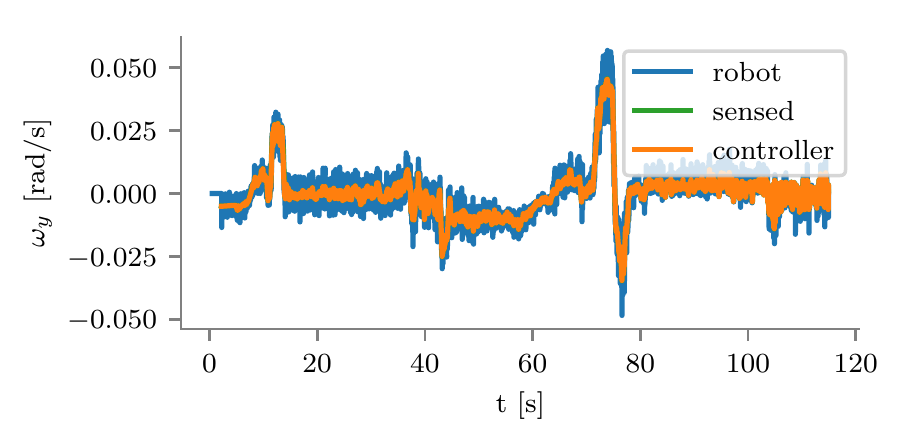}
    \caption{Complying with $\tau_y$}
  \end{subfigure}
  \begin{subfigure}[t]{\columnwidth}
    \centering
    \includegraphics[width=\columnwidth]{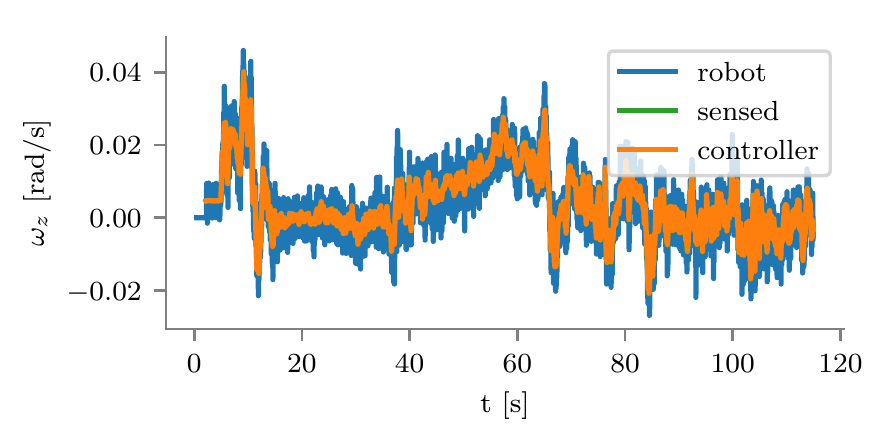}
    \caption{Complying with $\tau_z$}
  \end{subfigure}
  \caption{Sensor and controller angular speed (right and left hand
    side of \eqref{eq:examples:velocityconstraint:torque-comply}), and
    robot angular speeds of the end-effector. As the box constraint
    only considers the position of the end-effector, not the
    orientation, the box constraint does not affect the torque
    compliance.}
  \label{fig:examples:velocityconstraint:torque}
\end{figure}

In Fig.\ref{fig:examples:velocityconstraint:position} we see the
position of the end-effector over time.  In
Fig.\ref{fig:examples:velocityconstraint:forces} we see sensed speed,
controller speed and robot speed. The controller moves to track the
target function, resulting in compliance of the end-effector with
respect to the force. In
Fig.\ref{fig:examples:velocityconstraint:torque} we comply with
respect to the torques. From $t=36$s until $t=40$ we try to pull the
end-effector out of the box constraint. As the robot approaches the
box constraint, the compliance is reduced to zero. At $t=80$s we let
go of the end-effector and sensor bias moved it slowly to the bottom
and along the bottom of the box constraint. The noise is both a result
of the latency introduced by using ROS for real-time feedback control,
by the computation time of the controller, and by inherent noise in
the joint speed signal.

\section{Discussion}
Closed-loop inverse kinematics frameworks handle local problems rather
than global problems. When given desired positions far from the
current position, closed-loop inverse kinematics may succumb to local
minima. This means that they are mid-level controllers to which a
desired path may be supplied from a high-level path planner. The model
predictive controller formulation is an attempt at bridging the gap
between local and global planning. Proper design of cost and
constraint formulations to better achieve tasks may lead to better
handling of local minima. As of yet, the model predictive approach is
significantly slower than its reactive counterparts and further work
includes investigating sequential quadratic programming approaches
with warmstart as they may have shorter execution time than the
non-warmstarted interior point method of IPOPT. By using CasADi at the
core, CASCLIK can quickly benefit from any new solver implemented in
CasADi. The lack of convexity of the constraints along the prediction
horizon in the current formulation also suggests that further work
should be done to investigate the optimality and stability of the
approach.

CASCLIK is independent of the underlying representation used to define
kinematics. This allows for inspecting behavior of different
representations, but complicates programming for the user. A more
robust framework can be created by defining the underlying
representation. The choice of using arbitrary functions is a design
choice intended to allow a larger range of scenarios and systems to be
handled by CASCLIK as well as being easier to implement. Future work
includes examining other coordinate representations and constraints.

From the input experiments we see that the controllers have an
exponential convergence to the reference signal during positioning,
and classical compliance can be implemented in CASCLIK for the
controllers that accept velocity equality constraints. As sensors are
becoming cheaper and more available, it is important to allow for
arbitrary input signals. As the derivative of inputs are unknown, they
are considered to be zero. For input signals such as distance sensors
or force sensors, this is a false assumption and can lead to tracking
error for time varying sensor signals. Future work includes allowing
the user to provide input derivatives for CASCLIK. These may come from
speed observers, derivative filters, or any other sources the user
provides.

CASCLIK only considers control at the velocity setpoint level. This
stems from the main use-case for which the framework was intended,
industrial manipulators. In most cases, industrial manipulators only
provide joint position or joint velocity setpoints. However, the task
function approach allows for specifying control at the acceleration or
torque level \cite{Samson1991}. Extending CASCLIK to include
acceleration resolved controllers would allow specifying velocity
constraints that ensure convergence to the desired velocity. 

Although CasADi is intended for prototyping, C code generation is
possible. This could allow for specifying tasks and controllers with
CASCLIK before deploying the controller to an embedded system. For
control from an external computer the just-in-time compilation feature
of CasADi allows for defining quadratic problem or null-space based
controllers that can be run at $\leq$1 ms speeds, which can allow for
employing CASCLIK on real systems. However, for industrial use-cases
and real-time sensitive systems, the authors recommend eTaSL/eTC
\cite{Aertbelien2014}. Many robot systems have limits on acceleration
or jerk applied to the system, further work includes investigating
methods of implementing such constraints in CASCLIK.

From the examples, we see that the null-space approach has similar
behavior to the optimization approaches when handling a single set
constraint with very high gain. For multiple tasks, the null-space
projection operator will cause the set-based task priority framework
to behave differently from the optimization approaches. The
optimization approaches uses slack variables to handle multiple
tasks. This moves prioritization into the cost expression of the
optimization problem and does not allow for strict prioritization
between tasks, but tuning of the slack weight can be used to specify
different behavior of the tasks. As strict prioritization may be
desired in certain cases, future work includes creating a hybrid
approach that uses the optimization approach to define the desired
control variables for certain tasks, and the null-space projection to
ensure strict priority of other tasks.

The optimization approaches are closely related, and the complexity of
implementing them is similar. The null-space approach requires more
bookkeeping by the programmer but generally provides controllers with
shorter execution time. Generally the execution speeds are in the
order: null-space approach, QP approach, NLP approach, and MPC
approach. Each increasing by an order of magnitude in the sequence,
depending on the horizon length of the MPC. For rapid prototyping and
large set of tasks, the compilation time may also be of interest. As
the null-space approach compiles each separate mode, and there are
$2^{n_{set}}$ modes, its compilation time drastically increases with
multiple set constraints.

The nonlinear cost example shows that the NLP and MPC approach can
improve the manipulability in cases where the QP and null-space
approach did not. Although the MPC approach managed to find an
alternative configuration that increased the manipulability
significantly, the reactive NLP approach did not. As the
manipulability cost can be added to the QP formulation via Taylor
expansion of the cost as in \cite{Dufour2017}, the NLP approach may
not be as beneficial as initially expected.

\section{Conclusion}
In this paper, CASCLIK, a rapid prototyping framework for multiple
task-based closed-loop inverse kinematics controllers is presented. It
translates tasks into quadratic, nonlinear, or model predictive
optimization problems that can be solved with CasADi. Tasks are
formulated as constraints, and multiple tasks can be achieved
simultaneously.

CASCLIK also provides a CasADi based implementation of the set-based
task priority inverse kinematics framework. The paper includes a novel
multidimensional formulation of the \emph{in tangent cone} function
such that the set-based task priority framework can support
multidimensional set constraints. The results show that the
multidimensional set constraint formulation can give a better
representation of the desired behavior than when a multidimensional
set constraint is split into multiple one dimensional constraints.

\bibliography{library}{}

\begin{thebibliography}{10}
\providecommand{\url}[1]{#1}
\csname url@samestyle\endcsname
\providecommand{\newblock}{\relax}
\providecommand{\bibinfo}[2]{#2}
\providecommand{\BIBentrySTDinterwordspacing}{\spaceskip=0pt\relax}
\providecommand{\BIBentryALTinterwordstretchfactor}{4}
\providecommand{\BIBentryALTinterwordspacing}{\spaceskip=\fontdimen2\font plus
\BIBentryALTinterwordstretchfactor\fontdimen3\font minus
  \fontdimen4\font\relax}
\providecommand{\BIBforeignlanguage}[2]{{%
\expandafter\ifx\csname l@#1\endcsname\relax
\typeout{** WARNING: IEEEtran.bst: No hyphenation pattern has been}%
\typeout{** loaded for the language `#1'. Using the pattern for}%
\typeout{** the default language instead.}%
\else
\language=\csname l@#1\endcsname
\fi
#2}}
\providecommand{\BIBdecl}{\relax}
\BIBdecl

\bibitem{Sciavicco1986}
L.~Sciavicco and B.~Siciliano, ``{Coordinate Transformation: A Solution
  Algorithm for One Class of Robots},'' \emph{IEEE Transactions on Systems, Man
  and Cybernetics}, vol.~16, no.~4, pp. 550--559, 1986.

\bibitem{Samson1991}
C.~Samson, M.~{Le Borgne}, and B.~Espiau, \emph{{Robot Control: The Task
  Function Approach}}, 1st~ed.\hskip 1em plus 0.5em minus 0.4em\relax New York:
  Oxford University Press, 1991.

\bibitem{DeSchutter2007}
J.~{De Schutter}, T.~{De Laet}, J.~Rutgeerts, W.~Decr{\'{e}}, R.~Smits,
  E.~Aertbeli{\"{e}}n, K.~Claes, and H.~Bruyninckx, ``{Constraint-based Task
  Specification and Estimation for Sensor-Based Robot Systems in the Presence
  of Geometric Uncertainty},'' \emph{The International Journal of Robotics
  Research}, vol.~26, no.~5, pp. 433--455, may 2007.

\bibitem{Hanafusa1981}
H.~Hanafusa, T.~Yoshikawa, and Y.~Nakamura, ``{Analysis and Control of
  Articulated Robot Arms with Redundancy},'' \emph{IFAC Proceedings Volumes},
  vol.~14, no.~2, pp. 1927--1932, 1981.

\bibitem{Chiaverini1997}
S.~Chiaverini, ``{Singularity-robust task-priority redundancy resolution for
  real-time kinematic control of robot manipulators},'' \emph{IEEE Transactions
  on Robotics and Automation}, vol.~13, no.~3, pp. 398--410, 1997.

\bibitem{Moe2016}
S.~Moe, G.~Antonelli, A.~R. Teel, K.~Y. Pettersen, and J.~Schrimpf,
  ``{Set-Based Tasks within the Singularity-Robust Multiple Task-Priority
  Inverse Kinematics Framework: General Formulation, Stability Analysis, and
  Experimental Results},'' \emph{Frontiers in Robotics and AI}, vol.~3, no.
  April, pp. 1--18, 2016.

\bibitem{Aertbelien2014}
E.~Aertbelien and J.~{De Schutter}, ``{eTaSL/eTC: A constraint-based task
  specification language and robot controller using expression graphs},'' in
  \emph{IEEE International Conference on Intelligent Robots and Systems}.\hskip
  1em plus 0.5em minus 0.4em\relax IEEE, sep 2014, pp. 1540--1546.

\bibitem{Andersson2013b}
J.~Andersson, ``{A General-Purpose Software Framework for Dynamic
  Optimization},'' PhD thesis, Arenberg Doctoral School, KU Leuven, Belgium,
  2013.

\bibitem{Ferreau2014}
H.~J. Ferreau, C.~Kirches, A.~Potschka, H.~G. Bock, and M.~Diehl, ``{qpOASES: a
  parametric active-set algorithm for quadratic programming},''
  \emph{Mathematical Programming Computation}, vol.~6, no.~4, pp. 327--363,
  2014.

\bibitem{Wachter2006}
A.~W{\"{a}}chter and L.~T. Biegler, ``{On the implementation of an
  interior-point filter line-search algorithm for large-scale nonlinear
  programming},'' \emph{Mathematical Programming}, vol. 106, no.~1, pp. 25--57,
  mar 2006.

\bibitem{Arbo2018CASE}
M.~H. Arbo, Y.~Pane, E.~Aertbeli{\"{e}}n, and W.~Decre, ``{A System
  Architecture for Constraint-Based Programming of Robotics Assembly with CAD
  Information},'' in \emph{IEEE International Conference on Automation Science
  and Engineering}, 2018.

\bibitem{Mansard2009}
N.~Mansard, O.~Khatib, and A.~Kheddar, ``{A Unified Approach to Integrate
  Unilateral Constraints in the Stack of Tasks},'' \emph{IEEE Transactions on
  Robotics}, vol.~25, no.~3, pp. 670--685, jun 2009.

\bibitem{stack-of-tasks}
\BIBentryALTinterwordspacing
``{Stack-of-tasks}.'' [Online]. Available:
  \url{http://stack-of-tasks.github.io/}
\BIBentrySTDinterwordspacing

\bibitem{Escande2014}
A.~Escande, N.~Mansard, and P.~B. Wieber, ``{Hierarchical quadratic
  programming: Fast online humanoid-robot motion generation},''
  \emph{International Journal of Robotics Research}, vol.~33, no.~7, pp.
  1006--1028, 2014.

\bibitem{Smits2008}
R.~Smits, T.~{De Laet}, K.~Claes, H.~Bruyninckx, and J.~{De Schutter},
  ``{iTASC: a tool for multi-sensor integration in robot manipulation},'' in
  \emph{2008 IEEE International Conference on Multisensor Fusion and
  Integration for Intelligent Systems}.\hskip 1em plus 0.5em minus 0.4em\relax
  IEEE, aug 2008, pp. 426--433.

\bibitem{itasc-url}
\BIBentryALTinterwordspacing
``{OROCOS: iTaSC wiki}.'' [Online]. Available:
  \url{http://orocos.org/wiki/orocos/itasc-wiki/2-itasc-software}
\BIBentrySTDinterwordspacing

\bibitem{Bruyninckx2003}
H.~Bruyninckx, P.~Soetens, and B.~Koninckx, ``{The real-time motion control
  core of the Orocos project},'' \emph{2003 IEEE International Conference on
  Robotics and Automation (Cat. No.03CH37422)}, pp. 2766--2771, 2003.

\bibitem{rtt-url}
\BIBentryALTinterwordspacing
P.~Soetens, ``{RTT: Real-Time Toolkit}.'' [Online]. Available:
  \url{http://www.orocos.org/rtt}
\BIBentrySTDinterwordspacing

\bibitem{expressiongraphs-url}
\BIBentryALTinterwordspacing
E.~Aertbeli{\"{e}}n, ``expressiongraphs.'' [Online]. Available:
  \url{https://github.com/eaertbel/expressiongraph}
\BIBentrySTDinterwordspacing

\bibitem{kdl-url}
\BIBentryALTinterwordspacing
R.~Smits, ``{KDL: Kinematics and Dynamics Library}.'' [Online]. Available:
  \url{http://www.orocos.org/kdl}
\BIBentrySTDinterwordspacing

\bibitem{Somani2016}
N.~Somani, M.~Rickert, A.~Gaschler, C.~Cai, A.~Perzylo, and A.~Knoll, ``{Task
  level robot programming using prioritized non-linear inequality
  constraints},'' \emph{IEEE International Conference on Intelligent Robots and
  Systems}, vol. 2016-Novem, pp. 430--437, 2016.

\bibitem{Somani2017}
N.~Somani, M.~Rickert, and A.~Knoll, ``{An Exact Solver for Geometric
  Constraints With Inequalities},'' \emph{IEEE Robotics and Automation
  Letters}, vol.~2, no.~2, pp. 1148--1155, apr 2017.

\bibitem{Perzylo2016}
A.~Perzylo, N.~Somani, S.~Profanter, I.~Kessler, M.~Rickert, and A.~Knoll,
  ``{Intuitive instruction of industrial robots: Semantic process descriptions
  for small lot production},'' in \emph{2016 IEEE/RSJ International Conference
  on Intelligent Robots and Systems (IROS)}, vol. 2016-Novem.\hskip 1em plus
  0.5em minus 0.4em\relax IEEE, oct 2016, pp. 2293--2300.

\bibitem{Antonelli2009}
G.~Antonelli, G.~Indiveri, and S.~Chiaverini, ``{Prioritized closed-loop
  inverse kinematic algorithms for redundant robotic systems with velocity
  saturations},'' \emph{2009 IEEE/RSJ International Conference on Intelligent
  Robots and Systems, IROS 2009}, no.~3, pp. 5892--5897, 2009.

\bibitem{Arbo2018_syroco}
M.~H. Arbo and J.~T. Gravdahl, ``{Stability of the Tracking Problem with
  Task-Priority Inverse Kinematics},'' in \emph{12th IFAC Symposium on Robot
  Control SYROCO 2018}, vol.~51, no.~22, Budapest, 2018, pp. 121--125.

\bibitem{Boyd2006}
S.~Boyd and L.~Vandenberghe, \emph{{Convex Optimization}}.\hskip 1em plus 0.5em
  minus 0.4em\relax Cambridge: Cambridge University Press, 2004.

\bibitem{sumof1s_oeis}
\BIBentryALTinterwordspacing
``{A000788: Total number of 1's in binary expansions of 0, ..., n}.'' [Online].
  Available: \url{https://oeis.org/A000788}
\BIBentrySTDinterwordspacing

\bibitem{Colome2015}
A.~Colome and C.~Torras, ``{Closed-loop inverse kinematics for redundant
  robots: Comparative assessment and two enhancements},'' \emph{IEEE/ASME
  Transactions on Mechatronics}, vol.~20, no.~2, pp. 944--955, 2015.

\bibitem{mahaarbo-url}
\BIBentryALTinterwordspacing
M.~H. Arbo, ``{Mathias Hauan Arbo's Github page}.'' [Online]. Available:
  \url{https://github.com/mahaarbo/}
\BIBentrySTDinterwordspacing

\bibitem{Andersen2015}
T.~T. Andersen, ``{Optimizing the Universal Robots ROS driver.}'' Technical
  University of Denmark, Department of Electrical Engineering, Tech. Rep.,
  2015.

\bibitem{Figueredo2013}
L.~F. Figueredo, B.~V. Adorno, J.~Y. Ishihara, and G.~A. Borges, ``{Robust
  kinematic control of manipulator robots using dual quaternion
  representation},'' \emph{Proceedings - IEEE International Conference on
  Robotics and Automation}, pp. 1949--1955, 2013.

\bibitem{Larochelle2007}
P.~M. Larochelle, A.~P. Murray, and J.~Angeles, ``{A Distance Metric for Finite
  Sets of Rigid-Body Displacements via the Polar Decomposition},''
  \emph{Journal of Mechanical Design}, vol. 129, no.~8, p. 883, 2007.

\bibitem{Siciliano2008}
B.~Siciliano and O.~Khatib, \emph{{Springer Handbook of Robotics}}.\hskip 1em
  plus 0.5em minus 0.4em\relax Springer Science {\&} Business Media, 2008.

\bibitem{Wang2012}
X.~Wang, D.~Han, C.~Yu, and Z.~Zheng, ``{The geometric structure of unit dual
  quaternion with application in kinematic control},'' \emph{Journal of
  Mathematical Analysis and Applications}, vol. 389, no.~2, pp. 1352--1364,
  2012.

\bibitem{Dufour2017}
K.~Dufour and W.~Suleiman, ``{On integrating manipulability index into inverse
  kinematics solver},'' in \emph{IEEE International Conference on Intelligent
  Robots and Systems}, vol. 2017-Septe, 2017, pp. 6967--6972.

\end{thebibliography}
\bibliographystyle{IEEEtran}
\end{document}